%% file: main.tex
\pdfoutput=1

\documentclass{article}

\usepackage[final]{neurips_2022}

\usepackage[utf8]{inputenc} 
\usepackage[T1]{fontenc}    
\usepackage[hidelinks]{hyperref}       
\usepackage{url}            
\usepackage{booktabs}       
\usepackage{amsfonts}       
\usepackage{nicefrac}       
\usepackage{microtype}      
\usepackage[dvipsnames]{xcolor}         

\usepackage{pdfx}

\usepackage{amsmath,mathtools}
\usepackage[small]{caption}
\usepackage{graphicx,wrapfig}
\usepackage{enumitem}
\usepackage{algorithm,algpseudocode}
\usepackage{threeparttable}
\usepackage{comment}
\usepackage[utf8]{inputenc}

\setcitestyle{numbers,comma,open={[},close={]}}

\usepackage{titlesec}
\titlespacing*{\subsection}{0pt}{0.002\baselineskip}{0.001\baselineskip}
\titlespacing*{\section}{0pt}{0.004\baselineskip}{0.002\baselineskip}
\renewcommand{\footnotesize}{\fontsize{8pt}{10pt}\selectfont}

\hypersetup{colorlinks=true,linkcolor=orange,anchorcolor=black,citecolor=orange,filecolor=magenta,menucolor=black,runcolor=black,urlcolor=orange}

\algtext*{EndFor}
\algtext*{EndIf}

\newcommand{\argmin}{\operatornamewithlimits{arg\,min}}
\newcommand{\argmax}{\operatornamewithlimits{arg\,max}}
\newcommand{\argset}{\operatornamewithlimits{arg\,set}}
\DeclarePairedDelimiter\abs{\lvert}{\rvert}

\newcommand\mydots{\hbox to 1em{.\hss.\hss.}}

\title{
Assistive Teaching of Motor Control Tasks to Humans}

\author{%
  Megha Srivastava\\
  Stanford University\\
  \texttt{megha@cs.stanford.edu}\\
  \And
  Erdem B\i y\i k\\
  UC Berkeley\\
  \texttt{ebiyik@berkeley.edu}\\
  \And
  Suvir Mirchandani\\
  Stanford University\\
  \texttt{suvir@cs.stanford.edu}\\
  \And
  Noah D. Goodman\\
  Stanford University\\
  \texttt{ngoodman@stanford.edu}\\
  \And
  Dorsa Sadigh\\
  Stanford University \\ 
  \texttt{dorsa@cs.stanford.edu}\\
}

\begin{document}

\abovedisplayskip=3pt
\abovedisplayshortskip=3pt
\belowdisplayskip=2pt
\belowdisplayshortskip=2pt

\linepenalty=100
\maketitle
\input{intro}

\input{related-works}

\input{formulation}

\input{method}

\input{setup}

\input{experiments}

\input{discussion}

\section{Acknowledgements}
We thank all anonymous reviewers for their valuable feedback. We acknowledge support from Ford, AFOSR, and NSF Awards \#2218760, \#2132847, and \#2006388. Megha Srivastava was also supported by the NSF Graduate Research Fellowship Program under Grant No. DGE-1656518.
\\

\bibliographystyle{unsrtnat}
{
\small
\bibliography{refs}
}

\section*{Checklist}

The checklist follows the references.  Please
read the checklist guidelines carefully for information on how to answer these
questions.  For each question, change the default \answerTODO{} to \answerYes{},
\answerNo{}, or \answerNA{}.  You are strongly encouraged to include a {\bf
justification to your answer}, either by referencing the appropriate section of
your paper or providing a brief inline description.  For example:
\begin{itemize}
  \item Did you include the license to the code and datasets? \answerYes{Included in supplement.}
  \item Did you include the license to the code and datasets? \answerNo{The code and the data are proprietary.}
  \item Did you include the license to the code and datasets? \answerNA{}
\end{itemize}
Please do not modify the questions and only use the provided macros for your
answers.  Note that the Checklist section does not count towards the page
limit.  In your paper, please delete this instructions block and only keep the
Checklist section heading above along with the questions/answers below.

\begin{enumerate}

\item For all authors...
\begin{enumerate}
  \item Do the main claims made in the abstract and introduction accurately reflect the paper's contributions and scope?
    \answerYes{}
  \item Did you describe the limitations of your work?
    \answerYes{}
  \item Did you discuss any potential negative societal impacts of your work?
    \answerYes{We provide a detailed Ethics statement in the Appendix.}
  \item Have you read the ethics review guidelines and ensured that your paper conforms to them?
    \answerYes{}
\end{enumerate}

\item If you are including theoretical results...
\begin{enumerate}
  \item Did you state the full set of assumptions of all theoretical results?
    \answerNA{}
        \item Did you include complete proofs of all theoretical results?
    \answerNA{}
\end{enumerate}

\item If you ran experiments...
\begin{enumerate}
  \item Did you include the code, data, and instructions needed to reproduce the main experimental results (either in the supplemental material or as a URL)?
    \answerYes{In Appendix}
  \item Did you specify all the training details (e.g., data splits, hyperparameters, how they were chosen)?
    \answerYes{In Appendix}
        \item Did you report error bars (e.g., with respect to the random seed after running experiments multiple times)?
    \answerYes{}
        \item Did you include the total amount of compute and the type of resources used (e.g., type of GPUs, internal cluster, or cloud provider)?
    \answerYes{In Appendix}
\end{enumerate}

\item If you are using existing assets (e.g., code, data, models) or curating/releasing new assets...
\begin{enumerate}
  \item If your work uses existing assets, did you cite the creators?
    \answerYes
  \item Did you mention the license of the assets?
    \answerNA
  \item Did you include any new assets either in the supplemental material or as a URL?
    \answerNA
  \item Did you discuss whether and how consent was obtained from people whose data you're using/curating?
    \answerNA
  \item Did you discuss whether the data you are using/curating contains personally identifiable information or offensive content?
    \answerNA
\end{enumerate}

\item If you used crowdsourcing or conducted research with human subjects...
\begin{enumerate}
  \item Did you include the full text of instructions given to participants and screenshots, if applicable?
    \answerYes{In Appendix}
  \item Did you describe any potential participant risks, with links to Institutional Review Board (IRB) approvals, if applicable?
    \answerYes{Full details in Appendix.}
  \item Did you include the estimated hourly wage paid to participants and the total amount spent on participant compensation?
    \answerYes{In Appendix.}
\end{enumerate}

\end{enumerate}


\newpage
\appendix
\include{appendix}

\end{document}

%% file: intro.tex
\begin{abstract}
Recent works on shared autonomy and assistive-AI technologies, such as assistive robot teleoperation, seek to model and help human users with limited ability in a fixed task. However, these approaches often fail to account for humans' ability to adapt and eventually learn how to execute a control task themselves. Furthermore, in applications where it may be desirable for a human to intervene, these methods may inhibit their ability to learn how to succeed with full self-control. In this paper, we focus on the problem of \textit{assistive teaching} of motor control tasks such as parking a car or landing an aircraft. Despite their ubiquitous role in humans' daily activities and occupations, motor tasks are rarely taught in a uniform way due to their high complexity and variance. We propose an AI-assisted teaching algorithm that leverages skill discovery methods from reinforcement learning (RL) to (i) break down any motor control task into teachable skills, (ii) construct novel drill sequences, and (iii) individualize curricula to students with different capabilities. Through an extensive mix of synthetic and user studies on two motor control tasks---parking a car with a joystick and writing characters from the Balinese alphabet---we show that assisted teaching with skills improves student performance by around 40\% compared to practicing full trajectories without skills, and practicing with individualized drills can result in up to 25\% further improvement.\footnote{Our source code is available at \url{https://github.com/Stanford-ILIAD/teaching}.}
\end{abstract}

\section{Introduction}

Imagine a novice human is tasked with operating a machine with challenging or unfamiliar controls, such as landing an aircraft or parking an oversized vehicle with novel transmission. A fully autonomous form of assistance, such as a self-driving car, would aid this user by entirely replacing their control, seeking interventions only when necessary. Shared autonomous systems, such as those used in assistive robot teleoperation \cite{javdani2018shared, jeon2020shared, losey2021learning}, would seek to model the human's goals and capabilities, and provide assistance in the form of corrections \cite{reddy2018shared} or simplified control spaces \cite{losey2021learning, karamcheti2021lila}. However, neither form of assistance would actually help the user \textit{learn} how to successfully operate the machine, and may even serve as a crutch that prevents their ability to ever perform the task independently. 

Although humans can learn a variety of complex motor control tasks (e.g., driving a car, or operating surgical robots), they often rely on the assistance of specialized teachers. Receiving fine-grained instruction from these specialized teachers can often be non-uniform, costly, and limited by their availability. 
In this work, we are interested in a more accessible and efficient approach: we wish to develop an AI-assisted teaching algorithm capable of teaching motor control tasks in an individualized manner. While several works have leveraged AI techniques to aid instruction in traditional education tasks such as arithmetic \cite{rafferty2016faster, bassen2020reinforcement, poesia2021contrastive} and foreign language learning \cite{srivastava2021question, mu2021automatic}, motor control tasks introduce several unique challenges, such as the high complexity of the input and output space for motor control tasks (e.g., controlling a high degree of freedom robotic arm with a 6 degrees of freedom joystick or the continuous output space of trajectories in driving), as well as the need for generalization across different scenarios in these high-dimensional and complex spaces. 

To address these challenges, our insight is to utilize compositional motor skills for teaching: skills are more manageable to teach, and can be used to construct novel compositional drills for an individual. We take inspiration from how specialized human teachers teach motor control tasks. For example, a piano teacher may assign a student structured exercises (e.g., musical scales or arpeggios) to build up their general technique. Further, they may guide the student to break down complex musical phrases into more manageable chunks, repeatedly practice those, and then compose them together \cite{agay1982teaching}.

However, such fine-grained teaching often requires access to a specialized teacher capable of identifying skills and creating individualized drills based on a student's unique expertise. This is not only expensive, but can be highly suboptimal for many motor control tasks where a teacher may not have full visibility into the student's various inputs for a task (e.g. how much the student pushed the brake pedal), but only observability of the final behavior of the system (e.g. how close the vehicle is to the curbside), leading to less meaningful feedback. Can we leverage the rich action data provided by students learning motor control tasks to develop more efficient and reliable individualized instruction?

Our key insight is to leverage automated skill discovery methods from the RL literature to break down motor control tasks performed by an expert into \emph{teachable skills}.  
We then identify which of the expert's skills are skills that the student is struggling with, and construct \emph{novel drills} that focus on those particular skills to \emph{individualize} their curriculum. Our contributions include: 1) Developing an algorithm to identify which skills of a motor control task a student struggles with, 2) Automatically creating individualized drills based on students' expertise in various skills, and 3) Empirically validating the helpfulness of our AI-assisted teaching framework in two different motor control tasks: controlling a simulated vehicle to park with an unintuitive joystick, and writing Balinese words using a computer mouse or trackpad. Our results show that assisted teaching with skills improve student performance by around 40\% compared to practicing full trajectories without skills, and practicing with individualized drills can result in up to 25\% further improvement.

%% file: related-works.tex
\section{Related work}

\textbf{Skill discovery in Reinforcement Learning.}
A large body of work has studied how to leverage behavioral primitives, or skills in RL. Hierarchical RL approaches \citep{sutton1999between, barto2003recent} use skills as temporal abstractions to accelerate  learning \citep{mcgovern1998macro, parr1997reinforcement, dietterich2000hierarchical}.
However, hand-designing skills can be difficult, especially in continuous spaces. Several works proposed methods to automatically discover skills \cite{schaal2005learning, daniel2013autonomous, ranchod2015nonparametric, bacon2017option, bagaria2019option}. Works in unsupervised skill discovery derive task-agnostic skills directly from an environment, without a reward function \citep{gregor2016variational, florensa2017stochastic, achiam2018variational, eysenbach2019diversity, sharma2020dynamics, campos2020explore}. Skills can also be learned from a set of tasks \citep{frans2018meta} or demonstrations \citep{kipf2019compile, shankar2019discovering}; e.g., CompILE \citep{kipf2019compile} learns to segment and reconstruct given trajectories, using learned segment encodings to represent skills.
Rather than using discovered skills to aid the learning of automated agents (e.g., via RL or planning \citep{achiam2018variational, eysenbach2019diversity, sharma2020dynamics}), we investigate the novel application of discovered skills to teaching \emph{humans}.

\textbf{Assistive human-AI systems.}
Human-AI collaborative systems have the potential to augment a human’s individual capabilities. In the paradigm of \textit{shared autonomy} \citep{dragan2013policy}, an AI assistant modulates a user’s observation \citep{reddy2021assisted} or input on tasks such as robot teleoperation \citep{javdani2018shared, jeon2020shared, losey2021learning}. Unlike our work, these works generally assume the human’s policy is fixed, and do not directly aim for a human to learn from the AI agent \citep{reddy2018shared}. Bragg et al.~\cite{bragg2020fake} improve both collaborative performance and user learning in a shared autonomy framework, but they rely on simulated users in their evaluation. We present a technique to support human learning to improve individual proficiency, and rigorously evaluate it on real users.

\textbf{AI-assisted teaching.}
A distinct line of work has focused on developing algorithms for helping humans learn. \textit{Knowledge tracing} attempts to model a student’s knowledge state in terms of skills relevant to a set of tasks, yet have generally required manual skill annotations \citet{corbett1994knowledge} or large-scale student data \citep{piech2015deep}. \textit{Instructional sequencing} uses a model of student learning  to select a sequence of instructional activities to maximize learning, but these approaches have generally focused on generating curricula for domains like language learning, arithmetic, and logic \citep{doroudi2019where}. Similarly, \textit{machine teaching} \cite{zhu2015machine} explores the problem of selecting optimal training data for a student with a known model of learning, and has been effective in discrete domains with structured knowledge such as mathematics and visual classification \cite{mac2018teaching}. 

We build on the idea that a student’s proficiency can be represented in terms of different skills, and an AI assistant can adapt a curriculum accordingly. However, to the best of our knowledge, we are the first to explore how to teach motor control tasks informed by student actions. This setup has unique challenges: it is less obvious how to delineate skills, how to handle student feedback (in the form of continuous trajectories versus binary correct/incorrect responses), and how to measure a student’s proficiency at a skill. By leveraging advances from skill discovery in RL, we do not require expert annotation of skills nor large-scale student data.

\textbf{Motor learning in humans.}
\emph{Motor skill learning} in psychology refers to the acquisition of skilled actions as a result of practice or experience \citep{krahe1999motor, shumway1995theory}. Our approach is informed by several theories in motor learning and specifically the common insights that human motor control operates hierarchically, and that direct practice of individual skills generally aids skill learning \citep{krahe1999motor, rosenbaum2009human}.

%% file: formulation.tex
\section{Formulation}\label{sec:formulation}

In this section, we formalize the problem of \emph{Student-Aware AI-Assisted Teaching}. We go over preliminaries and formalize skill discovery, and then define the problem statement leading to our 3-step approach of teaching complex motor control tasks: (i) distilling tasks into underlying skills, (ii) identifying the student's proficiency, and (iii) creating novel and individualized curricula.

\subsection{Preliminaries}

We formalize the target motor control task as a standard Markov decision process (MDP) $\langle\mathcal{S}, \mathcal{A}, \rho, f, R, T\rangle$ with finite horizon $T$, where $\rho$ is the initial state distribution and $f$ is a deterministic transition function. In our formulation, $R$ is a random variable for reward functions such that each $r\sim R$ is a function $r:\mathcal{S}\times\mathcal{A}\to\mathbb{R}$. Given an initial state $s_0\sim\rho$ and a reward function $r\sim R$ (both of which are revealed at the beginning of the task), the goal is to maximize the cumulative reward over the horizon.
We call any $(s_0, r)$ pair a scenario $\xi$ drawn from a set of all possible scenarios $\Xi$. We assume access to a set of expert demonstrations, e.g., demonstrations drawn from a pre-trained RL agent that optimizes the cumulative reward for any given scenario. 

An inexperienced human would lack the expertise to perform optimally in a given scenario $\xi$. Our goal is to develop teaching algorithms that leverage expert demonstrations to improve student performance. For example, we want a human unfamiliar with Balinese to learn to write any Balinese word, which would correspond to one $r$.
For this, we develop an algorithm that uses skill discovery methods to automatically create individualized drills based on the discovered skills and the student performance. 

\textbf{Skill discovery.}
Humans often break down complex tasks into relevant skills (i.e., compositional action sequences that lead to high performance)---when learning how to write, we first master how to write letters, and then compose them to write words and sentences. 
We define a skill as a latent variable $m$ that lies in some finite discrete latent space $\mathcal{M}$. Each skill is an embedding that corresponds to an action sequence $\Lambda\!=\!(a_1,a_2,\dots,a_{\abs{\Lambda}})$ optimal at some state $s\!\in\!\mathcal{S}$ under some reward $r\!\sim\! R$:
\begin{align}
    \exists s\in\mathcal{S}, \exists r:\quad & P(R=r)>0, \quad \Lambda = (a^*_1,a^*_2,\dots,a^*_{\abs{\Lambda}}), \nonumber \\ & (a^*_1,a^*_2,\dots,a^*_{\abs{\Lambda}}, \dots) \in \argmax_{a_1,a_2,\dots,a_{\abs{\Lambda}},\dots} \left(r(s,a_1) + r(f(s,a_1),a_2) + \dots\right)\:,
\end{align}
and $\abs{\Lambda}\!\in\! [H_{\textrm{min}}, H_{\textrm{max}}]$, where $H_{\textrm{min}}$ and $H_{\textrm{max}} \ll T$ are hyperparameters.

In many tasks, it is difficult to enumerate all possible skills, and it is often not clear when a skill starts and ends in a trajectory. In this work, we use skill discovery algorithms to automatically extract skills relevant to a task, and then teach the student those generated skills. While our teaching method is agnostic to the skill discovery algorithm, we use CompILE \citep{kipf2019compile} as it learns skills directly from expert trajectories. Our formulation assumes access to expert trajectories, so it would be unnecessary to learn an expert policy in parallel with the skill discovery process,
as several other methods do. 

We use the trained CompILE as a function: $\textsc{SkillExtractor}(\tau)$ where $\tau$ is a trajectory. It outputs a sequence of skills $M^{\tau}\!=\!\left(m^{\tau}_1, m^{\tau}_2, \mydots\right)$ and boundaries $B^{\tau}\!=\!\left(b^{\tau}_0, b^{\tau}_1, \mydots, b^{\tau}_{N_\textrm{seg}}\right)$ such that $0 \!=\! b_0^\tau \!<\! b_1^\tau\! <\! \cdots\! <\! b_{N_\textrm{seg}}^\tau\! =\! T$, dividing $\tau$ into $N_\textrm{seg}$ \emph{segments}. Actions in $\tau$ between steps $[b^{\tau}_{j-1}, b^{\tau}_{j})$ correspond to skill $m^{\tau}_j\in \mathcal{M}$. Effectively, CompILE learns to divide trajectories into segments, and cluster segments with similar actions into skills. We refer to \citep{kipf2019compile} and the Appendix for more details. 
\color{black}

\subsection{Problem statement}
Having formalized $\textsc{SkillExtractor}$, which, given a trajectory $\tau$, outputs $M^\tau$ and $B^\tau$, we now present our problem statement for AI-assisted teaching of motor control tasks. 

\textbf{What should the student do?} Given a set of expert demonstrations in scenarios $\Xi^e\! \subseteq\! \Xi$, where each $\xi\!\in\!\Xi^e$ is sampled from $(\rho,R)$, we would ideally teach the student to eventually have a policy $\pi^*\!:\mathcal{S}\!\to\!\mathcal{A}$ that minimizes the difference between their behavior and the expert demonstrations:
\begin{align}
    \pi^* = \argmin_{\pi}\sum\nolimits_{\xi\in\Xi} P(\xi) \mathcal{L}(\tau^e_\xi, \tau^{\pi}_\xi) \approx \argmin_{\pi}\sum\nolimits_{\xi\in\Xi^e} \mathcal{L}(\tau^e_\xi, \tau^{\pi}_\xi)\:,
    \label{eq:teaching_objective}
\end{align}
where $\tau^{\pi}_\xi$ denotes the student trajectory performed by policy $\pi$ in scenario $\xi$. Similarly, $\tau^e_\xi$ denotes the expert trajectory in $\xi$.\footnote{For concision, we assume a single expert demonstration $\tau^e_\xi$ and at most a single student trajectory $\tau^\pi_\xi$ for any $\xi\in\Xi^e$. Without this assumption, we would take expectations over expert and student trajectories for each $\xi$.} $\mathcal{L}$ is a task-dependent loss function. While it is a straightforward goal, it is not clear how to teach a human to optimize this objective to achieve expert-like trajectories for any $\xi$.

\begin{wrapfigure}{R}{0.45\textwidth}
    \vspace{-15px}
	\includegraphics[width=\linewidth]{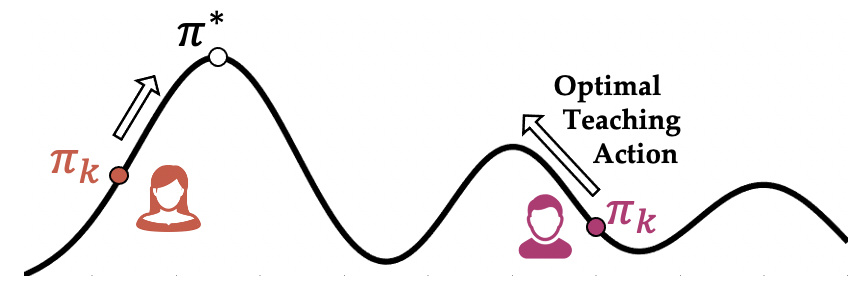}
	\centering
	\vspace{-15px}
	\caption{An optimal teacher should select a teaching action (e.g. which practice scenario to provide) based on each student's individual policy to help guide them towards the expert policy $\pi^*$.
	}
	\vspace{-10px}
	\label{fig:teaching_landscape}
\end{wrapfigure}
\textbf{What should the teacher do?} We can formulate the teaching problem as a partially observable MDP (POMDP), where the state of this POMDP is the student's policy (initially $\pi_0$), and the teacher's actions guide the student to the policy $\pi^*$. The teacher's observations would be student behavior expressed by their trajectories, or $\tau^{\pi_k}_\xi$ for some $\xi$ for state $\pi_k$ at teaching round $k$. The teacher can then select actions (e.g. practice scenarios) in this POMDP. A teaching action at round $k$ moves the student from $\pi_k$ to $\pi_{k+1}$, giving the teacher a reward of $\sum_{\xi\in\Xi}P(\xi)\left(\mathcal{L}(\tau^e_\xi, \tau^{\pi_k}_\xi) - \mathcal{L}(\tau^e_\xi, \tau^{\pi_{k+1}}_\xi)\right)$, i.e., the decrease in the loss value according to Eq.~\eqref{eq:teaching_objective}. Thus, the teacher should take the optimal teaching actions that maximize its cumulative reward over teaching rounds, whose global optimum would move the student to $\pi^*$ (see Fig.~\ref{fig:teaching_landscape}).

Unfortunately, we lack  computational models of human learning necessary for making the POMDP's state transition function (i.e., how a student updates its policy from $\pi_k$ to $\pi_{k+1}$ for any teaching action the teacher could take) tractable for motor control tasks. Recent works exploring data-driven frameworks for knowledge tracing \cite{piech2015deep} and RL-based education \cite{bassen2020reinforcement}  usually require large amounts of data for simpler problems, and are thus not scalable to motor control tasks with large trajectory spaces.

\textbf{Approach overview.} Our insight is that once we identify important skills of a motor control task, an AI assisted teaching framework can efficiently leverage the skills to teach the task.
The \textsc{SkillExtractor} function allows us to identify the skills used in a given trajectory. Applying it to all expert demonstrations in scenarios $\Xi^e\subseteq\Xi$ gives us a set of skills relevant in the task. Say we label the expert demonstration $\tau^e_\xi$ for each scenario $\xi\in\Xi^e$ with skills and their intervals: $(M^e_\xi, B^e_\xi) = \textsc{SkillExtractor}(\tau^e_\xi)$.
We would then want the student to perform each relevant skill as close as possible to the expert to solve the following optimization:
\begin{align}
    \pi^* \!=\! \argmin_{\pi}\sum\nolimits_{\xi \in \Xi} P(\xi) \sum\nolimits_{m \in M_\xi^e} \mathcal{L}_m(\tau^e_\xi, \tau^\pi_\xi) \!\approx\! \argmin_{\pi}\sum\nolimits_{\xi \in \Xi^e}\! \sum\nolimits_{m \in M_\xi^e} \mathcal{L}_m(\tau^e_\xi, \tau^\pi_\xi)\:,
    \label{eq:skill_objective}
\end{align}
where $\mathcal{L}_m$ is some loss function that compares expert and student trajectories in terms of the skill $m$. The intuition behind this objective is two-fold: (i) we want the student to perform the same set of skills as the expert for any $\xi\in\Xi$ (this is why we sum over $m\in M^e_\xi$), and (ii) we want them to perform each skill as close as possible to how the expert performs it. The underlying assumption is that student will gain expertise if they perform the same skills in the most similar way to the expert.

Overall, \eqref{eq:skill_objective} is a simpler objective as the space of action sequences for each skill is much smaller than the space of trajectories or scenarios. We can now try to teach a small number of skills to the student instead of unrealistically trying to match them with the expert in the large trajectory space. 

Only teaching popular skills of a task may suffer from the problem that the student does not learn how to compose different skills to achieve the target task. We thus propose to create novel \emph{drills}, which consist of repetitions of multiple skills in a single trajectory. Drills help the students (i) learn how to connect action sequences for different skills, and (ii) develop muscle memory, which is crucial in motor control tasks, for the action sequences of each skill.
The generated drills can also be personalized based on what skills each student is struggling with. For instance, as in Fig.~\ref{fig:teaching_landscape}, two different students (in green and blue) might have different policies with different skill sets. Therefore, a teacher should generate \emph{individualized drills} informed by the student's initial level of expertise.

To efficiently solve \eqref{eq:skill_objective}, we propose our student-aware teaching approach that follows three steps: \textbf{1) Diverse scenario selection:} Identify scenarios that yield a high diversity of popular skills when an expert controls the system, \textbf{2) Expertise estimation:} Have the student control the system under these scenarios to estimate their expertise in different skills, and \textbf{3) Individualized drill generation:} Based on the estimates, create individualized drills to have the student practice the skills that they are struggling with.
In the subsequent section, we describe our procedure for each of these steps.

%% file: method.tex
\section{Student-aware assisted teaching}
\label{sec:method}

\begin{figure}[t]
	\includegraphics[width=\textwidth]{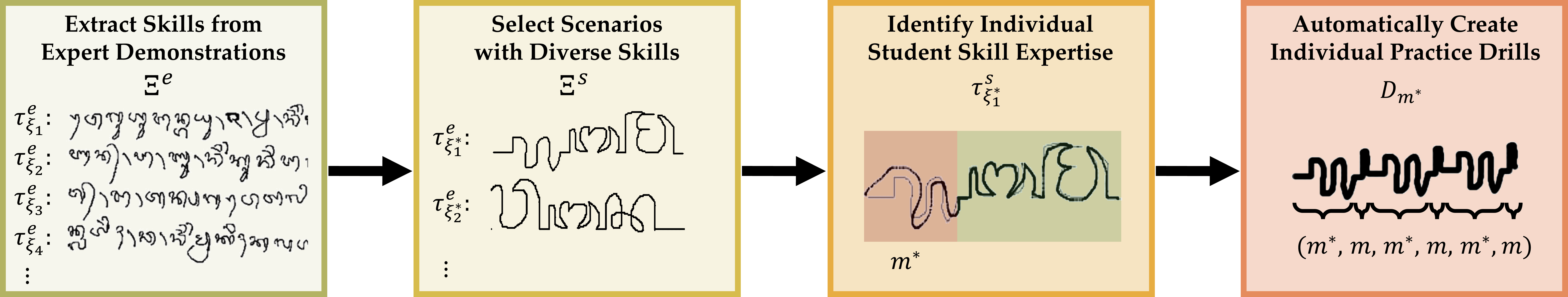}
	\centering
	\vspace{-15px}
	\caption{\textbf{Overview of our approach.} We train a CompILE model over expert demonstrations (e.g., handwritten Balinese text) for our \textsc{SkillExtractor} function, which we use to select a diverse set of scenarios (e.g. words) covering many skills. After a student provides trajectories for these scenarios, we use \textsc{SkillExtractor} again to identify their individual expertise for each skill, which informs the set of drills we provide them to practice.}
	\vspace{-20px}
	\label{fig:overview}
\end{figure}

\begin{wrapfigure}{R}{.49\linewidth}
\vspace{-45px}
\begin{minipage}{\linewidth}
\begin{algorithm}[H]
\caption{Diverse Scenario Selection}
\label{alg:diverse_scenario_selection}
\begin{algorithmic}[1]
    \Require{Skill labels $M^e_\xi$ for each $\xi\in\Xi^e$}
    \Require{Number of scenarios to be selected $N^s$}
    \State{$\Xi^s, M_{\textrm{covered}}\gets \emptyset, \emptyset$}
    \For{$i=1,2,\dots,N^s$}
        \State{$\xi^* \gets \argmax_{\xi\in\Xi^e\setminus\Xi^s}\abs{M_{\textrm{covered}} \cup M^e_{\xi}}$}
        \State{$\Xi^s \gets \Xi^s \cup \{\xi^*\}$}
        \State{$M_{\textrm{covered}} \gets M_{\textrm{covered}} \cup M^e_{\xi}$}
    \EndFor
    \State\Return{$\Xi^s$}
\end{algorithmic}
\end{algorithm}
\end{minipage}
\vspace{-15px}
\end{wrapfigure}
\textbf{1) Diverse scenario selection.} We select diverse scenarios that cover many different skills in order flexibly estimate students' expertise and individualize our teaching strategy. For example, in English, we may have expert demonstrations of writing the words (scenarios) ``expert'', ``person'', ``skills'', from which we would select ``person'' to cover more diverse letter strokes (skills). Formally, we treat the problem of choosing scenarios with diverse skills as a maximum set coverage problem, which we solve using a simple greedy algorithm (see Algorithm \ref{alg:diverse_scenario_selection}), outputting a set of scenarios that led the expert to demonstrate a variety of skills.

\begin{wrapfigure}{R}{.49\linewidth}
\vspace{-25px}
\begin{minipage}{\linewidth}
\begin{algorithm}[H]
\caption{Individual Expertise Identification}
\label{alg:expertise_identification}
\begin{algorithmic}[1]
    \Require Selected scenarios $\Xi^s\subseteq\Xi^e$
    \Require Skill labels $M^e_\xi$ for each $\xi\in\Xi^e$
    \Require Student demonstration $\tau^s_\xi$ for each $\xi\!\in\!\Xi^s$
    \State $E_m \gets 0$ for $\forall m\in\mathcal{M}$
    \For{$\xi \in \Xi^s$}
        \State $M^s_\xi, B^s_\xi \gets \textsc{SkillExtractor}(\tau^s_\xi)$
        \State $j\gets 0$
        \For{$m\in M^e_\xi$}
            \If{$m\not\in M^s_\xi$}
                \State{$E_m \gets E_m + r_{\xi}(\tau_\xi^s)/j$}
            \EndIf
            \State{$j \gets j + 1$}
        \EndFor
    \EndFor
    \State\Return $E$
\end{algorithmic}
\end{algorithm}
\end{minipage}
\vspace{-15px}
\end{wrapfigure}

\textbf{2) Identifying individual expertise in skills.} To efficiently teach complex tasks, we seek to estimate student's expertise in different skills, such as identifying which strokes they struggle with when writing Balinese words.
Requiring a diverse set of skills, the scenarios $\Xi^s$ we select enable us to infer student's expertise efficiently: we ask the student to perform the task in those scenarios, and based on their data, construct a labels vector $E$ that quantifies the student's expertise $E_m$ for each skill $m\!\in\!\mathcal{M}$.

The discrepancy between skills used by the student and the expert provides a useful signal to estimate the student's expertise.
Intuitively, we expect the student to demonstrate similar skills to the expert if they are proficient in those skills, while, otherwise, they might rely on a simpler or incorrect set of skills.
For example, a student driver who lacks the expertise of reverse driving might attempt to park a vehicle by only driving forward.
We therefore utilize the expert demonstrations in the same scenarios the student performed the task by extracting the skills the student used via $\textsc{SkillExtractor}$ and comparing them with the expert skills in those scenarios. The student's expertise in a skill $m$ is then: $E_m \!=\! -\sum_{\xi\in\Xi^s}\!\Delta_m\!(\tau^e_\xi, \tau^s_\xi)$,
where $\tau^s_\xi$ is the student trajectory in $\xi$, and $\Delta_m$ is a function that computes the discrepancy between the two trajectories in terms of $m$.

In our implementation, if a skill is used only by the expert, we assume the student did not attempt or failed that skill. We increase the discrepancy based on how far in the trajectory they failed. We assign higher penalty to failures in the beginning of the scenario for two reasons: 1) Skills relevant early in the task are often more important: other skills would not even be needed without first achieving these skills. 2) Failing at early skills might mean the student never had a chance or a need to perform the later skills due to compounding errors,
so our uncertainty about the later skills is higher. Thus, we decide to be more conservative about the later skills.
Specifically, we use:
\begin{align}
    \Delta_m(\tau^e_\xi, \tau^s_\xi) = \begin{cases}
    -r_\xi(\tau^s_\xi)/j & \textrm{ if } m\not\in M^s_\xi \textrm{ and } m^{\tau^e_\xi}_j = m,\\
    0 & \textrm{ otherwise.}
    \end{cases}
    \label{eq:discrepancy}
\end{align}
Here $r_\xi$ is the reward function under scenario $\xi$, which we assume to be non-positive.\footnote{This is a mild assumption, since we can always subtract a constant offset from any finite reward function.} We increase the discrepancy if a skill is only performed by the expert, and we discount this value over time ($j$ in Eq.~\eqref{eq:discrepancy}). We also scale discrepancy with $r_\xi(\tau^s_\xi)$ so that student trajectories with lower reward will be penalized more in terms of expertise. Alg.~\ref{alg:expertise_identification} presents the pseudocode for this step. In applications where the reward function is not readily available, one could design the skill-discrepancy function $\Delta_m$ independent of the rewards by using a divergence metric between expert and student trajectories.

\textbf{3) Creating individualized drills.} Referring to Eq.~\eqref{eq:skill_objective}, a student should master the skills they struggle with to get better at the overall task. Therefore, the teacher should first identify those skills, and then have the student practice them. For example, if a teacher thinks a student who attempted to write the word ``person'' struggled with writing the letter ``p'', they could provide the word ``puppet'' as a \emph{drill} for the student. Using the expertise vector $E$ for a student, which is the output of Alg.~\ref{alg:expertise_identification}, we can select the skills $m^*$ with low $E_{m^*}$ as the skills that the student needs to practice.

\begin{wrapfigure}{R}{.50\linewidth}
\vspace{-20px}
\begin{minipage}{\linewidth}
\begin{algorithm}[H]
\caption{Individualized Drill Creation}
\label{alg:drill_creation}
\begin{algorithmic}[1]
    \Require Individual expertise vector $E$
    \Require Number of skills to create drills for $N_\textrm{target}$
    \Require Hyperparameters $n, N_{\textrm{drills}}, N_{\textrm{rep}}$
    \Require Expert demonstration $\tau^e_\xi$ for each $\xi\in\Xi^e$
    \Require Skill labels $M^e_{\xi}$ for each $\xi\in\Xi^e$
    \State Compute $f(x)$ for all $n$-grams $x$\Comment{Eq.~\eqref{eq:frequency}}
    \For{$m^* \gets \argset_{N_{\textrm{target}}} \min_m E_m$}\Comment{see $\dagger$}
        \State $X\gets \argset_{N_{\textrm{drills}}} \max_{x\owns m^*} f(x)$\Comment{see $\dagger$}
        \State $D_{m^*} \gets \bigcup_{x \in X} \bigoplus_{i=1}^{N_{\textrm{rep}}} \bigoplus_{m \in x} \tau^e_{m}$\Comment{see $\ddagger$}
    \EndFor
    \State \Return $D$
\end{algorithmic}
\end{algorithm}
\begin{tablenotes}[para,flushleft]
    \vspace{-13pt}
    \begin{small}
	$\dagger$ $\argset_n \min_m E_m$ returns the set of $n$ distinct indices that individually minimize $E_m$.\\
	$\ddagger$ $\tau^e_{m}$ is the segment of an expert demonstration from interval $[b^{\tau^e}_{j-1}, b^{\tau^e}_{j})$ such that $m^{\tau^e}_j= m$. The operator $\oplus$ concatenates the actions of trajectory segments.
	\end{small}
\end{tablenotes}
\end{minipage}
\vspace{-15px}
\end{wrapfigure}
Given a skill $m^*$, the student should ideally practice it in the context of skills that frequently go together with $m^*$, rather than in isolation or with arbitrary skills. 
Our approach uses the expert demonstrations to identify frequent skill sequences containing skill $m^*$. We create \emph{drills} consisting of these sequences that enable us to have the student practice (i) the target skill $m^*$, and (ii) how skill $m^*$ connects with other skills that co-exist in various scenarios. Each drill repeats such a sequence $N_{\textrm{rep}}$ times. 
A natural approach is to iterate over all $n$-grams, borrowing terminology from natural language processing, of skills from the expert demonstrations. Within each $n$-gram, we check if $m^*$ exists to identify the skills that often co-occur with $m^*$. Mathematically, \emph{frequency} of an $n$-gram of skills $x$ is (where $\mathbb{I}$ is an indicator for the particular $n$-gram $x$):
\begin{align}
    f(x) = \sum\nolimits_{\xi\in\Xi^e}\sum\nolimits_{i=1}^{\abs{M_\xi^e}-n+1} \mathbb{I}\left(x = (m^{\tau_\xi^e}_i, m^{\tau_\xi^e}_{i+1},\ldots, m^{\tau_\xi^e}_{i+n-1})\right)\:,
    \label{eq:frequency}
\end{align}
We can then ask the student to practice the $N_{\textrm{drills}}$ $n$-grams involving the target skill $m^*$ with the highest frequency.\color{black}

Next, we formalize a drill as a repetition of the skills in the $n$-grams.
This ensures the student gets enough practice of skill $m^*$ with the most common skill sequence containing $m^*$. We then project this constructed skill sequence back to the trajectory space by again using the expert demonstrations. While different segments of the expert demonstrations may correspond to the same skill, we randomly select one of those segments to project back to the trajectory space.\footnote{We set the initial state of each drill as the initial state of the first trajectory segment in the drill. Since the transition function $f$ is deterministic, the initial state and action sequence specify a full trajectory.}
Alg.~\ref{alg:drill_creation} presents the pseudocode. It returns a set of drills $D_{m^*}$ for every target skill $m^*$. Though these drills can be used in several ways, e.g., explaining them to a student, here we let the student observe and then imitate each drill trajectory.

%% file: setup.tex
\section{Experiments}
\textbf{Environments.} We consider two motor control tasks: parking a car in simulation with a joystick controller (\textsc{Parking}), and writing Balinese characters (\textsc{Writing}) using a computer mouse \footnote{Note that the skills extracted are specific to operating a joystick/pen with a computer mouse. }. 

\textsc{Writing}: We introduce a novel task of writing Balinese character sequences\footnote{We select Balinese because, in comparison to other scripts available in the Omniglot dataset, the writing trajectories were consistent across users. 
 Assisted teaching when there are multiple ways of performing a task is an important future direction.} from  the Omniglot dataset \cite{lake2015human}, which contains action sequences for 1623 characters spanning 50 alphabets provided by crowdworkers from Amazon Mechanical Turk. We sample 5 Balinese characters, introduce a connector character ``-'', and create goal sequences of up to 8 characters. An agent in this environment operates in 2D ($x$,$y$) continuous state and action space, and receives reward corresponding to the degree of overlap between the agent's and the gold trajectory. We train CompILE with 1000 random  sequences based on trajectories provided by human contributors to the Omniglot dataset.

\textsc{Parking}: We use the Parking environment from  HighwayEnv \cite{highway-env}, a goal-based continuous control task where an agent must park a car in a specified parking spot by controlling its heading and acceleration (2D action space). The  6D state space corresponds to the car's position, velocity, and heading and at each time step, and the agent receives a reward representing how close the current state is to the goal state. Scenarios differ across initial heading angles and goal parking spots, resulting in a diverse set of skills needed. We train CompILE with rollouts across random scenarios from an optimal RL agent trained for $10^6$ epochs using the StableBaselines \cite{stable-baselines} implementation of Soft Actor-Critic. 

For both tasks, reward is negative with an optimal value of 0, but reported with a positive offset. \color{black} Further details, as well as necessary pre-processing steps for both environments and all hyperparameters used for training CompILE models as part of \textsc{SkillExtractor}, are in the Appendix. 

\textbf{Drill Creation.} 
We create one drill for each latent skill identified by \textsc{SkillExtractor} for both \textsc{Parking} ($n=3, N_{\textrm{rep}}=1, N_{\textrm{target}}=7, N_{\textrm{drills}}=1$) and \textsc{Writing} ($n=2, N_{\textrm{rep}}=3, N_{\textrm{target}}=8, N_{\textrm{drills}}=1$) tasks. Fig.~\ref{fig:pie} shows examples of these drills.

\textbf{User Study.} We design web applications for our user study for both environments. For \textsc{Parking}, students see the environment rendered on the interface, and control the car's steering and acceleration with a 2D joystick. For \textsc{Writing}, students use their computer mouse or trackpad to write on an HTML5 canvas, with the goal sequence of Balinese characters displayed underneath. Students in both environments follow a sequence of pre-test scenarios, practice sessions (including skills or drill-based practice), and evaluation scenarios. We recruit students through the Prolific platform, and include further information about pay rate, average experiment times, and task interfaces in the Appendix.

%% file: experiments.tex
\subsection{Results}

\textbf{Does individualization help synthetic students?} To demonstrate our full approach, we first show results for synthetic ``students'' in the \textsc{Parking} environment. We train two different synthetic students using standard behavior cloning on goal-conditioned $(s,a)$ tuples from expert demonstrations:

\begin{enumerate}[nosep,leftmargin=*]
    \item \emph{Half-trained:} We train a 4-layer feed-forward neural network for only 50 epochs, resulting in around 50\%  increase in Mean Squared Error than a student trained for 400 epochs. 
    \item  \emph{Reversing Difficulty:} We train a 4-layer feed-forward neural network for 400 epochs, but with only 20\% of the data containing reverse acceleration actions, to simulate difficulty with reversing. 
\end{enumerate}

Next, following the approach described in Sec.~\ref{sec:formulation}, we create a pool of 25 scenarios covering a diverse set of skills returned by our trained \textsc{SkillExtractor}. We then collect rollouts from each synthetic student across all scenarios, use Alg.~\ref{alg:expertise_identification} to identify which 3 skills each synthetic student has least expertise in, and select drills from our offline drill dataset that target these skills. Using an equal dataset size, we compare and report average student reward across 15 different evaluation sets, where the student is fine-tuned on $(s,a)$ sampled from these individualized drills (\textbf{Ind. Drills}) vs. random drills (\textbf{Drills}) vs. entire $(s,a)$ trajectories from the original expert demonstrations (\textbf{Full Trajectory}). 

\begin{wrapfigure}{R}{0.35\textwidth}
    \vspace{-23px}
	\includegraphics[width=\linewidth]{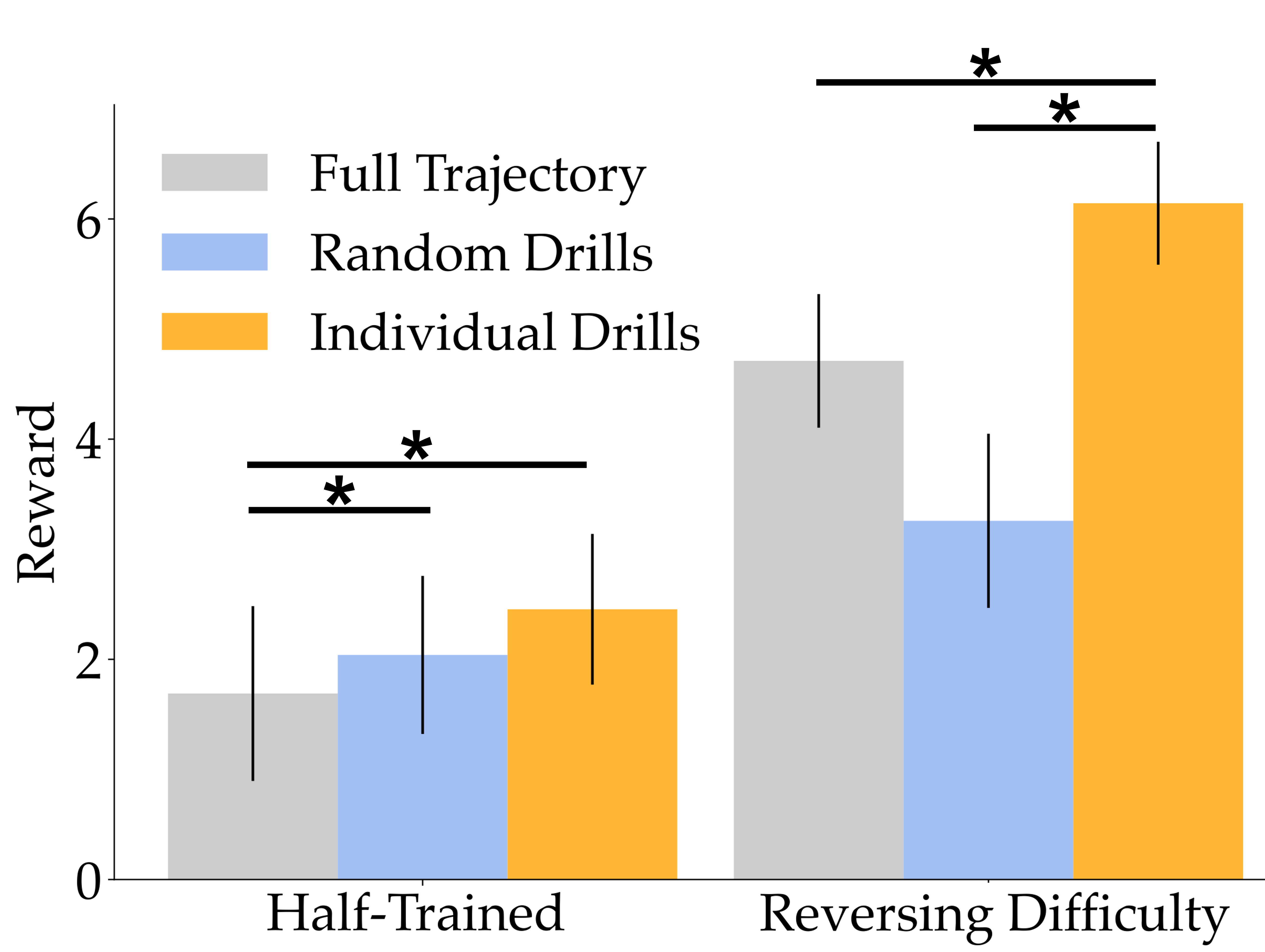}
	\centering
	\vspace{-17px}
	\caption{The effect of \textbf{Ind. Drills} and random \textbf{Drills} for synthetic students.}
	\vspace*{-10px}
	\label{fig:synthetic}
\end{wrapfigure}

Our results in Fig.~\ref{fig:synthetic} show that for the synthetic Half-trained IL Student, fine-tuning on data from both \textbf{Ind. Drills} and randomly selected \textbf{Drills} significantly outperforms the \textbf{Full Trajectory} setting ($p<.05$, Wilcoxon signed-rank test), while the difference between \textbf{Ind. Drills} and \textbf{Drills} is not significant. This may be due to a half-trained IL agent not having  low ``expertise'' in any particular region, but more generally benefiting from training further on good quality data. Interestingly, for the  Student with Reverse Action Difficulty, \textbf{Ind. Drills} significantly outperforms \textit{both} randomly selected \textbf{Drills} and \textbf{Full Trajectory} settings, but randomly selecting \textbf{Drills} slightly underperforms \textbf{Full Trajectory}, perhaps due to several drills not containing as many reverse actions as in the full-trajectory data. These results suggest that while drills can benefit both types of synthetic students in general, due to being composed of very common skill sequences, the synthetic student designed to lack a certain kind of ``skill'' is more likely to strongly benefit from individualization. This presents a promising signal that our CompILE-based \textsc{SkillExtractor} learns skills that align with our intuitions for what skills are necessary for the \textsc{Parking} task. However, our eventual goal is to teach \textit{humans} motor control tasks, so we next test the usefulness of our approach with human students.

\textbf{Does assisted skill-based practice improve human learning outcomes?}
Before constructing and individualizing drills, it is important to verify whether the skills returned by \textsc{SkillExtractor}---the building blocks of our approach---do indeed benefit student performance, as well as in comparison to simpler heuristics. For example, CompILE, as well as other skill discovery methods, take as input the expected number of skills $K$ per expert demonstration. A simpler \textbf{Time Heuristic}  approach could instead be to assume that skills are temporally ordered and have equal length,  and just split each expert demonstration into $K$ equally-sized segments. If such a heuristic sufficed, then it may not be necessary to leverage more complex skill discovery algorithms.We therefore run user studies for both \textsc{Parking} and \textsc{Writing} environments comparing three different settings:
\begin{enumerate}[nosep,leftmargin=*]
	\item \textbf{Skills:} Each user practices trajectories corresponding to the 3 most common skills returned by \textsc{SkillExtractor} based on CompILE parameterized with $K$, for 3 sessions each. 
	\item \textbf{Time Heuristic:} We temporally split each trajectory into $K$ intervals of equal length, and each user practices 3 different intervals for 3 sessions each. 
	\item \textbf{Full Trajectory:} Each user practices 3 different full trajectories of an expert demonstration (e.g. parking a car from the start state in \textsc{Parking}, an entire character sequence in \textsc{Writing}).
\end{enumerate}
We set $K=3$ for \textsc{Parking} and $K=8$ for \textsc{Writing}. For fair comparison, the number of time-steps users are able to practice in the full trajectory and time-heuristic modes is roughly equivalent. Because our users may differ widely in prior experience (e.g. a video game player may be more familiar with the joystick in \textsc{Parking}), we measure and report \textit{Reward Improvement}, or the difference in average reward across 5 random evaluation scenarios and 2 random pre-test scenarios. 

\begin{wrapfigure}{R}{.55\textwidth}
    \vspace{-15px}
	\includegraphics[width=\linewidth]{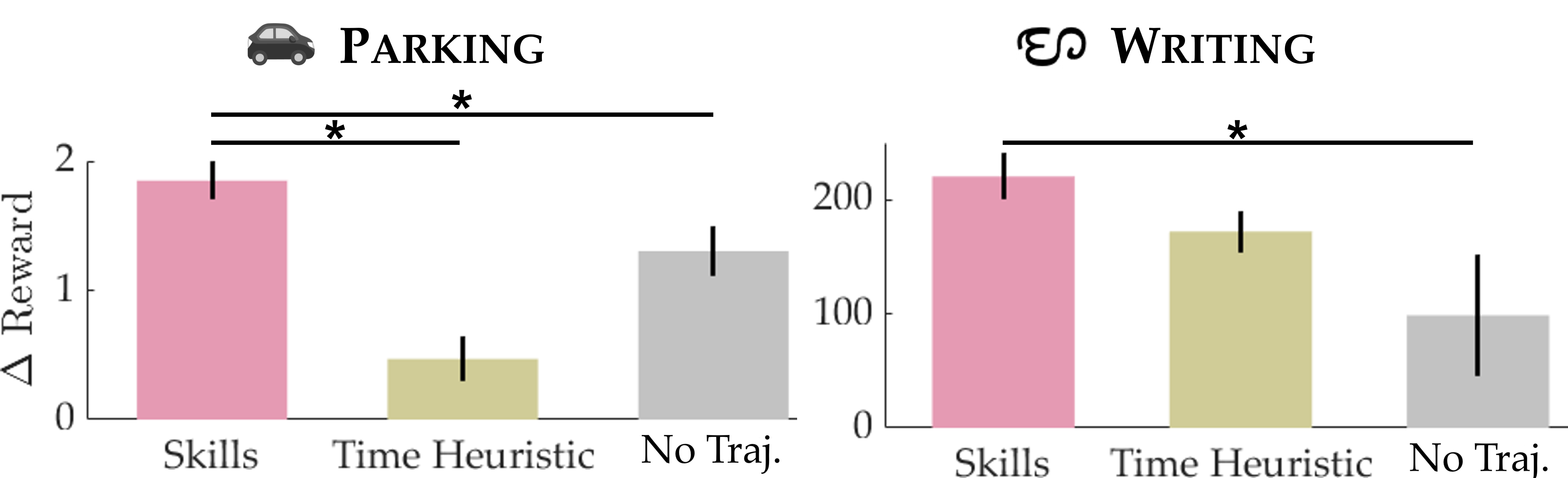}
	\centering
	\vspace{-15px}
	\caption{Students receive significantly higher reward improvement when practicing with the  \textbf{Skills} setting than \textbf{Full Trajectory}. A simple \textbf{Time Heuristic} results in inconsistent performance across environments, demonstrating the necessity for a more reliable way to identifying skills for teaching.}
	\vspace{-15px}
	\label{fig:skill}
\end{wrapfigure}

Our results in Fig.~\ref{fig:skill} show significantly ($p < .05$) higher reward improvement for students in both \textsc{Writing} ($N$ = 25) and \textsc{Parking} ($N$ = 20) in the \textbf{Skills} setting vs. \textbf{Full Trajectory} setting. This suggests that teaching students the necessary skills for a motor control task is extremely beneficial to improve learning. However, comparison between the \textbf{Skills} and \textbf{Time Heuristic} is interestingly inconsistent between environments, where \textbf{Time Heuristic} significantly leads to worse student learning outcomes in \textsc{Parking}, but not for \textsc{Writing}. One possible explanation might be that individual skills in \textsc{Writing}, such as characters or certain curves, might be semantically intuitive on their own, and therefore even if \textbf{Time Heuristic} presents different characters as the same overall ``skill'', there is less confusion for users. On the other hand, because of the challenging dynamics in \textsc{Parking}, a user may rely more on practicing skill trajectories that have clearly similar actions. Because heuristics are often environment-specific and complex, we view these results as an encouraging sign of the suitability of intelligent skill-discovery methods for teaching humans.  

Finally, we ask all participants in a post-study survey \emph{``What else would have been helpful for you to learn how to write the characters or how to park a car?''}.
One user for the \textsc{Writing} task who was assigned practice \emph{without} skills responded  \emph{``Separating the characters''}. Responses from users who were assigned skill-based practice included \emph{``Practicing a string of 2 characters''} and \emph{``More time and more repetition''}. These responses motivate our next set of experiments on the effect of drills.

\begin{wrapfigure}{R}{0.49\textwidth}
    \vspace*{-15px}
	\includegraphics[width=\linewidth]{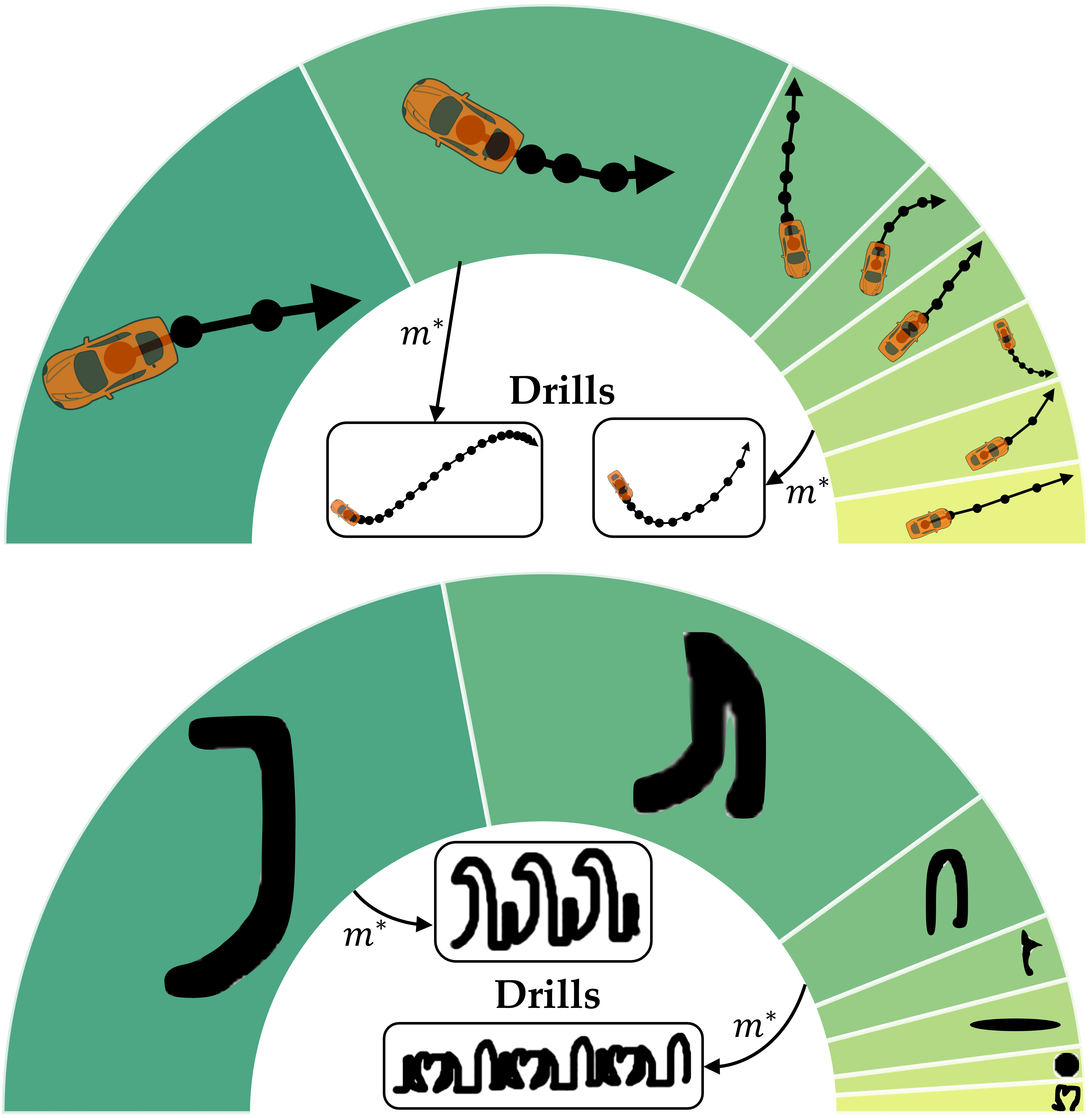}
	\centering
	\vspace{-15px}
	\caption{ Alg.~\ref{alg:expertise_identification} identifies a wide range of skills as low expertise across students for both environments. Reverse actions are identified as the most common skills for students to improve on in \textsc{Parking}, while common curved-shapes are identified for \textsc{Writing}.}
	\label{fig:pie}
	\vspace*{-15px}
\end{wrapfigure}

\textbf{Does practice with novel drill sequences outperform skill-based practice?}
Though our previous results demonstrate skill-focused practice leads to stronger learning outcomes in both environments, we hypothesize it is equally important to learn how to combine skills that often co-occur via \emph{drills}. For example, a qualitative look at the skills in Fig.~\ref{fig:pie} suggests some of the discovered skills consist of few actions, such as a short dot in \textsc{Writing} or only a few time steps in \textsc{Parking}. On their own these may not be useful skills to practice, but may be important to teach in the context of other skills. 

\begin{wrapfigure}{R}{.55\textwidth}
    \vspace{-17px}
	\includegraphics[width=\linewidth]{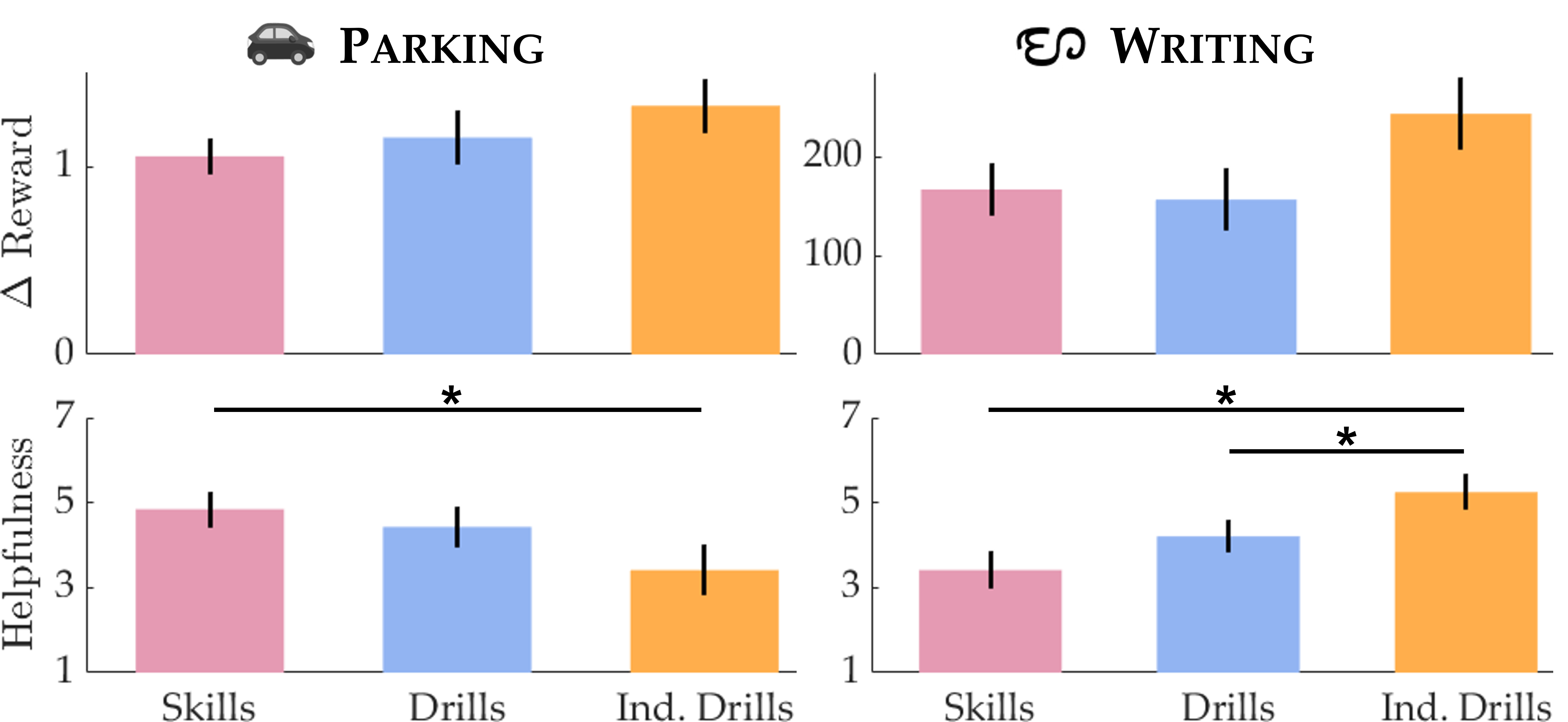}
	\centering
	\vspace{-15px}
	\caption{Students marginally improve reward more when practicing with \textbf{Ind. Drills} than \textbf{Skills} for both tasks. In \textsc{Writing}, students significantly prefer \textbf{Ind. Drills} over both random \textbf{Drills} and \textbf{Skills}, yet this does not hold in \textsc{Parking}, where targeted practice of reversing may frustrate students.}
	\vspace{-10px}
	\label{fig:drills}
\end{wrapfigure}
To compare with our individualization method, we create a fixed set of 5 pre-test scenarios covering a diverse set of skills returned by \textsc{SkillExtractor} for both environments. We select 2 random drills, and compare the change in student reward when practicing with drills (\textbf{Drills}) vs. practicing the same underlying skill sequences used to composed the drills as separate practice sessions (\textbf{Skills}).
Overall, as shown in Fig.~\ref{fig:drills}, we find no statistically significant difference in reward improvement when students practice repetitive drill sequences composed of multiple skills (Blue bar) vs. skills in separation (Pink bar). Possible explanations include the increased number of practice sessions in the \textbf{Skills} setting (because we keep the number of total time-steps equivalent across settings) mitigating benefits of repetition or practicing skill transitions in \textbf{Drills}. However, we next ask whether we may observe benefit from drills if we were to individualize them to students.

\textbf{Does individualizing drills improve student learning outcomes?}
For each student, we run a CompILE-based \textsc{SkillExtractor} immediately after receiving all pre-test student trajectories. Fig.~\ref{fig:pie} demonstrates the top 2 skills identified as ``low expertise'' across all students for both \textsc{Parking} ($N=20$) and \textsc{Writing} ($N=25$) environments. \textsc{SkillExtractor} combined with Alg.~\ref{alg:expertise_identification} is indeed able to identify a wide set of low-expertise skills in student trajectories, ranging from more popular skills (e.g. ``horn'' shape in \textsc{Writing}) to rarely difficult skills (e.g. steering the car forward in a straight line in \textsc{Parking}). Importantly, this is despite the likely strong differences between human student and expert trajectories, emphasizing the viability of our approach to teaching humans.

Next, we select 2 unique drills targeting each identified skill, and provide them to students for practice before measuring reward improvement on an evaluation set. Fig.~\ref{fig:drills} shows individualized drills (\textbf{Ind. Drills}) lead to stronger reward improvements for students than \textbf{Skills} with marginal significance ($p\!\approx\!.06$ \textsc{Writing}, $p\!\approx\!.09$ \textsc{Parking}), supporting our hypothesis that individualizing drills leads to better learning outcomes. We visualize all student trajectories before and after practicing \textbf{Ind. Drills} in the Appendix.  

Finally, we ask students to rate the helpfulness of practice sessions from 1-7 (1 = least helpful), and show in Fig.~\ref{fig:drills} for \textsc{Writing}, students strongly preferred practice sessions from \textbf{Ind. Drills} over \textbf{Drills} and \textbf{Skills}. However, despite the higher average reward improvement for \textbf{Ind. Drills} in \textsc{Parking}, students who participated in \textbf{Skills} practice rated the helpfulness of their sessions significantly higher. Though there may exist many explanations for this result, including student preference for shorter sequences, we observe in evaluation rounds that users who received \textbf{Ind. Drills} were around twice as likely to try \textbf{reversing} into a parking spot than other students (\textbf{53\% vs. 27\%}). This is likely due to our individualization method providing drills with reversing skills, which were more common targeted skills (see Fig.~\ref{fig:pie}). Crucially, in these scenarios \emph{the expert action is also to reverse}, suggesting that while individualization may create a more frustrating learning experience, it may still help students achieve stronger learning outcomes by teaching more challenging skills.

%% file: discussion.tex
\section{Discussion}
Our work is a first step towards building AI-assisted systems to more efficiently teach humans motor control tasks. However, there exist a few important limitations, which we now outline below.
\subsection{Extracting ``Human Teachable'' Skills}

A possible limitation of our work is whether skills extracted via \textsc{SkillExtractor} are suitable for teaching human learners. While one benefit of our  framework is its agnosticism to the \textit{type} of the expert, we could imagine that using an automated RL-trained expert, for example, might result in teaching skills that are unsuitable for human learners. Addressing this requires stronger cognitive models of human motor learning in order to iterate different types of skill decomposition and optimize for some notion of ``teachability.''
We could consider adapting existing methods used to model learning over time in other education domains (e.g. mathematics), such as Deep Knowledge Tracing (\citep{piech2015deep}) and Item Response Theory, which typically represent questions as discrete items or with simple features, and student responses as binary (correct or incorrect). Unfortunately, adapting such methods to handle the rich information in student trajectories is quite unstable, likely due to high dimension and scale (e.g. > 1K timesteps). Furthermore, student response variation for the same scenario is a lot higher for motor tasks than in other tasks such as in mathematics, where within a short period a student likely answers the same question the same way. Finally, such approaches require large amounts of student data (on the scale of thousands). Further discussion, as well as results demonstrating the unreliability of asking human experts to identify skills instead, is in Appendix Section \ref{sec:humanexpert}.

\subsection{Accounting for Student \& Expert Multimodality}
One important aspect of many motor control tasks is multimodality - there may exist many optimal ways to complete the task. Although one can naively extend our approach to handle this by collecting demonstrations from a diverse set of experts and identifying which expert to ``match'' a student with, understanding the different control preferences present in a motor task is complex. For example, in writing, while different stroke orders for the same character clearly constitute different modes, there exist more subtle differences, such as the degree of rotation used, or the sharpness of the trajectory. These may occur due to physical preferences or conditions such as arthritis, but are hard to automatically distinguish as separate modes versus noise. Cleanly defining the distinct modes of a control task likely requires strong domain knowledge combined with large amounts of data across a diverse population. Furthermore, inferring which mode to teach a novice student is challenging, as pre-test trajectories only provide a limited amount of information. In traditional education settings, we often ask students questions such as whether one is left or right handed, and use that to inform how we teach. However, there may exist more obscure forms of preferences that require longer interaction with a student to elicit, and identifying such preferences is an important direction for future work.   
\subsection{Potential Failure Cases}
Although we provide a general framework for assistive teaching of any arbitrary motor control task, including those that lack many expert instructors like teleoperating a novel surgical robot system, there are some settings that may be incompatible with our approach. These include situations where important skills do not appear in observed trajectories (e.g. the positioning of a hand on a ball before pitching in baseball), or there exists extreme stochasticity such that the same actions lead to drastically different outcomes, resulting in unintuitive dynamics for learners. Furthermore, in some tasks learning may not be feasible just from practice, but requires stronger forms of teacher-student interaction, such as physical guidance. We believe understanding the limits of demonstration-based teaching for motor tasks necessitate further insights from several areas within the broader NeurIPS research community, including cognitive science, reinforcement learning, and social aspects of machine learning.

%% file: appendix.tex
\textbf{Note}: Additional visualizations of our experiments can be found here: \url{https://sites.google.com/view/assistive-teaching-2022/home}

\section{Broader Impact \& Ethics Statement}
AI-assisted teaching of motor control tasks can provide significant benefits such as more reliable teaching to individual students with different abilities (e.g. by leveraging more granular information about student actions), adaptability to any type of motor task or expert agent, and improved safety by reducing burden on human teachers for safety-critical tasks. 
However, we emphasize that our approach is solely meant to \textit{assist} human teaching, as there exist many important aspects of human instruction that would be challenging to replace, including providing inspiration and motivation, in depth knowledge of human physical limitations, and an awareness of the broader context of a specific motor control task. Further risks of our approach, and avenues to address them, include: 
\begin{itemize}[nosep,leftmargin=*]
    \item \textbf{Bias of the expert agent.} The suitability of the skills we use for teaching relies on how diverse the set of demonstrations from an expert is. For example, if a writing task only contained demonstrations from right-handed experts, certain action sequences that may be harmful for left-handed students' learning may be chosen as skills. Understanding how CompILE performs over a mix of expert types, and how to enable more complex adaptation to a specific student's needs at the skill-identification step itself are important future directions.  
    \item \textbf{Over-reliance on the expert.} Our work currently assumes that in order to learn a task, the student should practice drills built from how an expert performs the task. However, a student should also be encouraged to learn when it may be appropriate to differ from the expert's actions if it helps the student learn better. This requires knowledge of how individual skills serve the ultimate task's goal (e.g. understanding why we first turn to enter a parking spot), which future work on incorporating interpretability methods and natural language instructions into our approach can address. 
    \item \textbf{Student physical constraints during learning.} In many tasks, certain action sequences may physically be easier for an expert to perform than a student, and may perhaps even be dangerous for a student to practice without building up necessary techniques. This can be addressed by leveraging more complex hierarchical approaches to skill discovery (e.g. which skills should be mastered before attempting others) and incorporating knowledge of human physical constraints (e.g. degree of feasible wrist rotation). 
    Another interesting direction for future work is to compare skills identified at different levels of expertise, and ascertain whether skills identified from a ``medium-level student'' may actually be easier, and less physically demanding, to \textit{teach} with than those identified from an expert.   
\end{itemize}

\section{Notation Glossary}
For convenience, we provide a glossary of all mathematical notation used in our framework. 
 \begin{table}[H]
\begin{center}
\begin{tabular}{ |c|c| } 
 \hline
 \textbf{Term} & \textbf{Meaning}  \\ 
 \hline
 $\Xi^s$/$\Xi^e$ & Student/Expert Scenarios \\ 
 \hline
 $\tau^s_\xi$/$\tau^e_\xi$ & Student/Expert Trajectories for scenario $\xi$ \\ 
 \hline
 $M^e_\xi$ & Set of skill labels corresponding to an Expert $e$'s trajectory for scenario $\xi$ \\ \hline
 $E$ & Expertise vector for a given student \\
 \hline
 $b^{\tau}$ & Boundary of a skill subsequence in a trajectory $\tau$ corresponding to a particular timestep  \\
 \hline
 $\tau^e_{m}$ & Segment of an expert $e$'s trajectory from interval $[b^{\tau^e}_{j-1}, b^{\tau^e}_{j})$ such that $m^{\tau^e}_j= m$ \\
 \hline
 $\oplus$ & Concatenation operator that stitches together action sequences when creating drills \\
 \hline
\end{tabular}
\end{center}
 \end{table}
\newpage

\section{Student Trajectories Before and After Practicing w/ Individualized Drills}

\begin{figure}[H]
    \centering
    \includegraphics[width=0.8\linewidth]{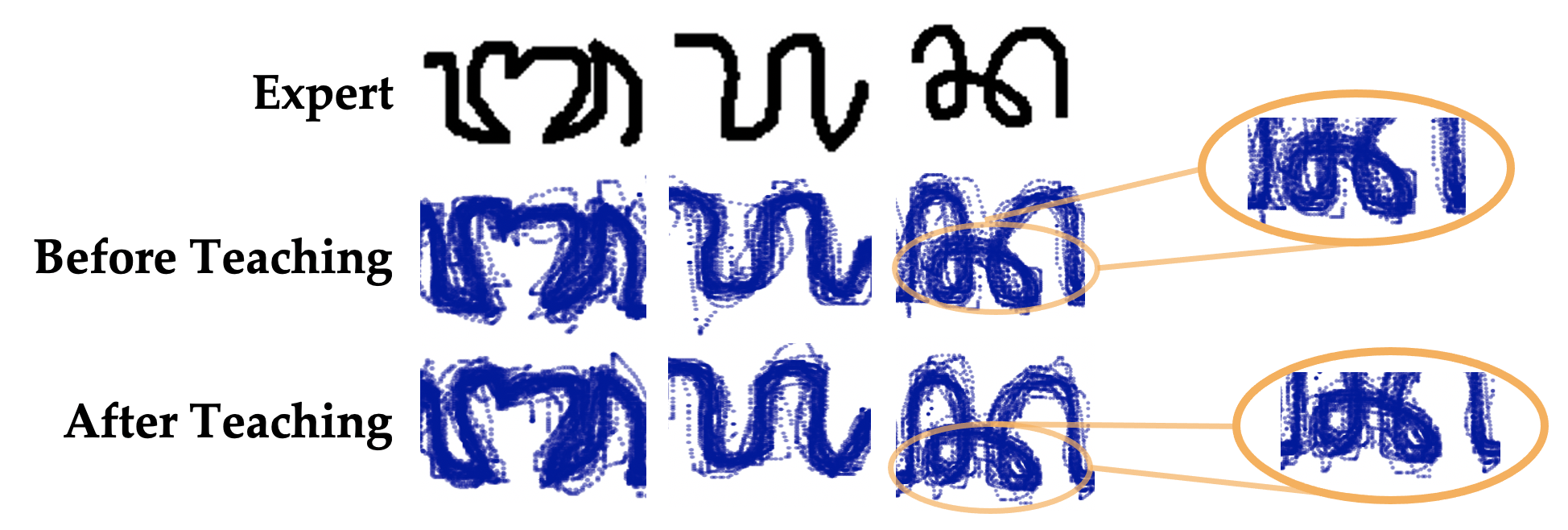}
    \caption{Overlay of all student trajectories of 3 Balinese characters for the \textsc{Writing} task before teaching with individualized drills and after. For the rightmost character (``na'' in Balinese), students learn to draw a smaller second loop at a lower angle, following the expert more closely. For other characters, we find that students exhibit less noise, and off-character strokes more closely following the original character.}
    \label{fig:student_examples}
\end{figure}

We note that while Fig. \ref{fig:student_examples} compares student trajectories before and after teaching for one particular set of Balinese characters, the reported values are averaged across all pre-test and evaluation rounds. Furthermore, common failure modes in the pre-test rounds before teaching that are not captured by this visualization include students giving up early and letting go of the mouse, and students re-tracing their characters. 

\section{Comparison Between Student and Expert CompILE Outputs}
\begin{figure}[H]
    \centering
    \includegraphics[width=0.6\linewidth]{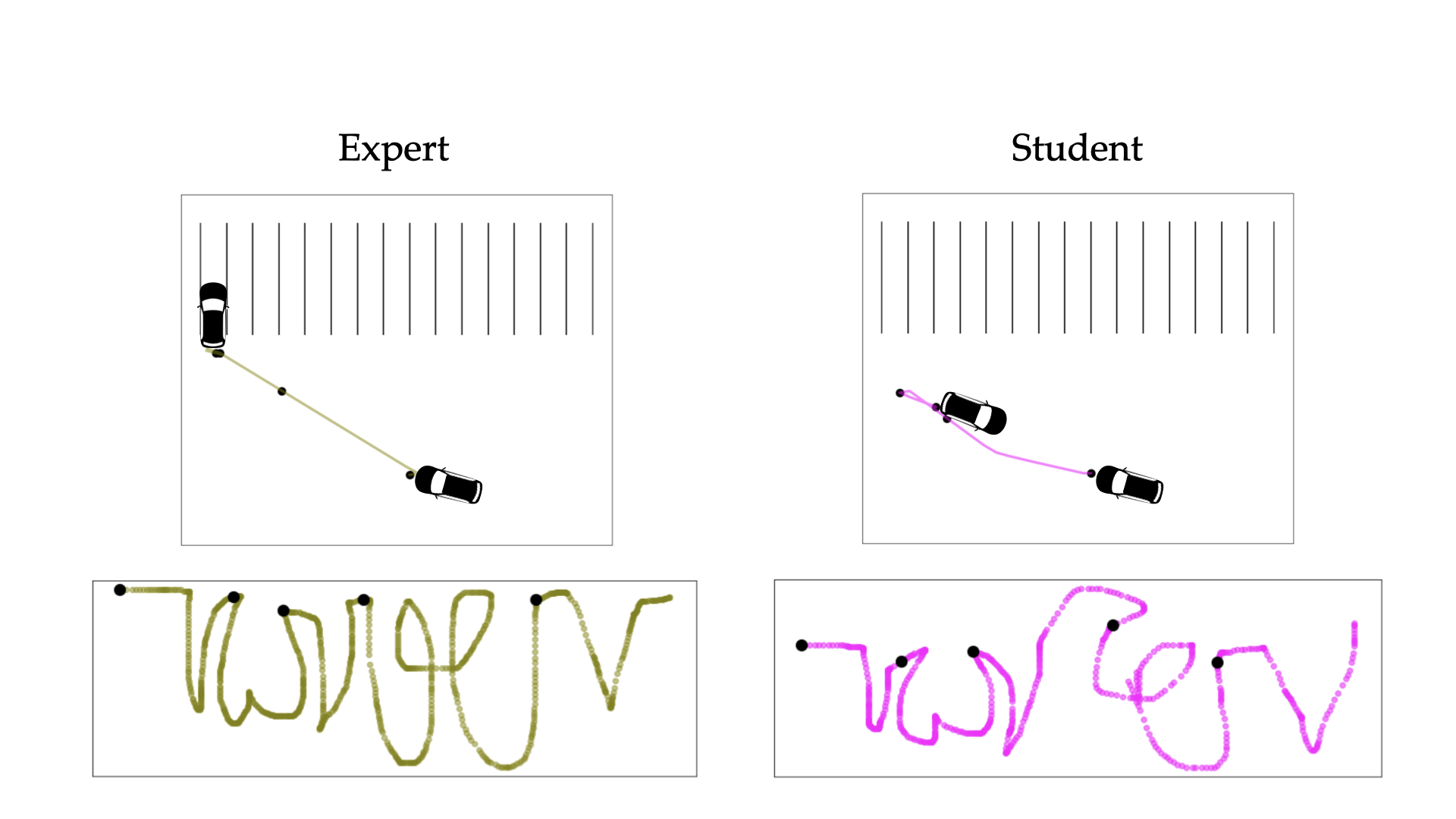}
    \caption{\textsc{SkillExtractor} outputs for Expert (left) and Student (right) trajectories for both the Parking (top) and Writing (bottom) tasks. Black dots represent skill boundaries identified by CompILE.}
    \label{fig:expert_vs_student}
\end{figure}
Comparing \textsc{SkillExtractor} output boundaries on trajectories from both our experts and students for both tasks, we can see that CompILE is able to segment both types of trajectories, but the noise in student trajectories leads to a failure of identifying the necessary skills for the task. We leverage this information to identify which skills students are struggling with.  For example, in \textsc{Parking}, the student is clearly unable to park the car, but the initial movement towards the upper left is segmented similarly to the expert, so the student will largely be penalized for later parts of the trajectory.

\section{Challenges in Human Expert Skill Identification}
\label{sec:humanexpert}
To address the challenges of extracting ``human teachable'' skills discussed above, one may consider using human experts as part of the \textsc{SkillExtractor} function. However, the key idea behind our work is that people who are experts at performing a task may not be expert at teaching it, and may struggle to identify skills consistently. 

\begin{figure}[H]
    \centering
    \includegraphics[width=0.8\linewidth]{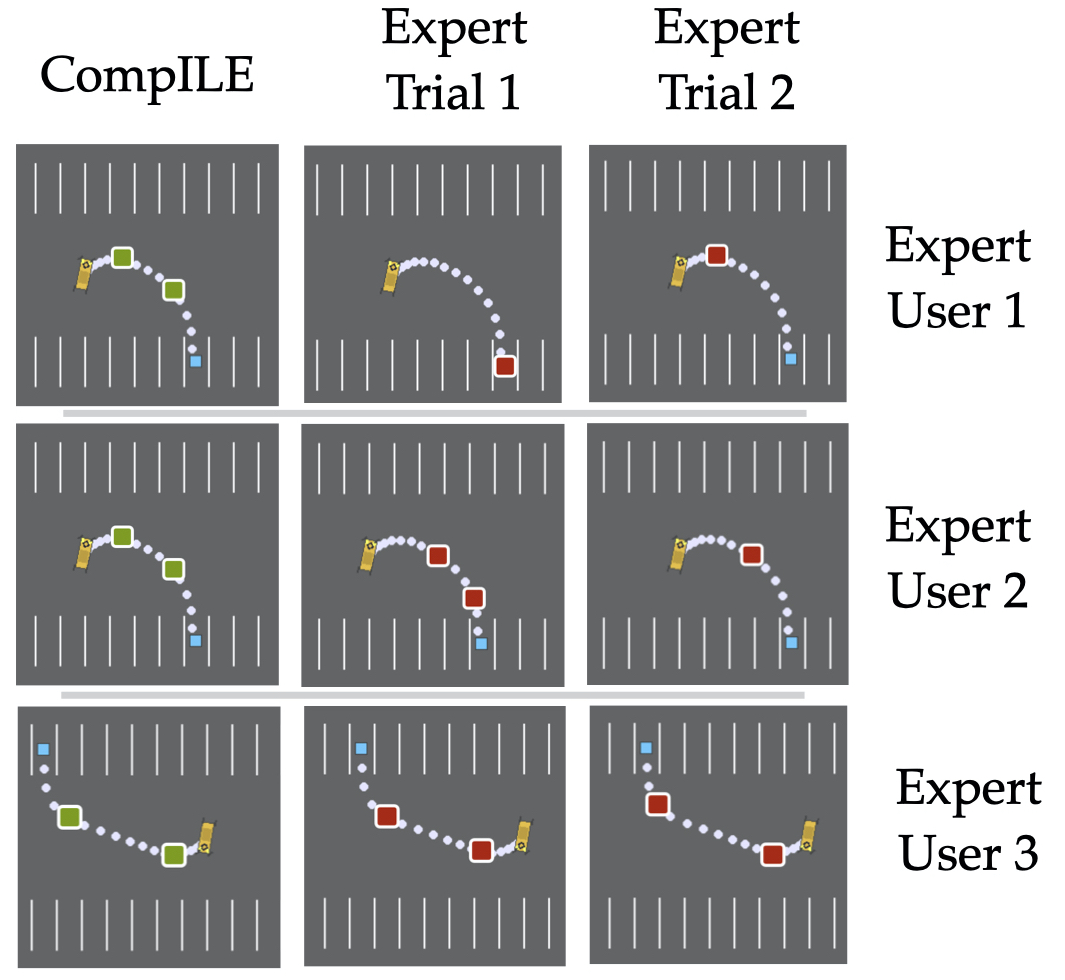}
    \caption{Skill segmentations from our CompILE-based SkillExtractor and 3 different human experts at the \textsc{Parking} task. For a given trajectory, experts provided 2 segment boundary annotations (columns), with each row corresponding to a different expert. Expert Users 1 and 2 show high variability among their own segmentations, while Expert User 3 is more consistent and provides skill segmentations similar to the output of CompILE.}
    \label{fig:expert_compare}
\end{figure}

To observe this clearly, we asked 3 different experts to annotate 10 successful trajectories of the \textsc{Parking} task with boundaries corresponding to skills under unlimited time. The set of 10 trajectories only contained 3 unique demonstrations, allowing us to measure whether experts were consistent when providing skill annotations for the \textit{same} expert demonstration. Fig. \ref{fig:expert_compare}
shows that even for the same trajectory, the same expert user may provide a completely different skill segmentation, and even identify a different number of skills. Even Expert 3, the most consistent expert across all duplicate trajectories, showed slight variation across trials - however, we note that CompILE's segmentations closely matched theirs. Moroever, by design, CompILE returns the exact same segmentations for the same action sequence. Overall, our user study results showed that human expert provided skills are often unreliable, leading to  skill sequences with high variations that would be challenging to teach in a uniform way.  

Finally, we note that a key aspect of our approach is scaling the ability to identify skills within a wide range of student trajectories for individualization. This is an even higher burden for human experts, who need to identify skills over trajectories that may widely differ from each other and the way the expert knows how to complete the task.

Therefore, in this work we attempted to incoporporate preliminary notions of  ``human teachability" when selecting between hyperparameter settings  of our CompILE-based \textsc{SkillExtractor}. Specifically, we filtered out skills corresponding to trajectories below a minimum length (due to human perceptual limits), and then chose the parameters that corresponded to the set of skills with highest entropy, with the intuition that a sufficiently diverse set of scenarios for a task would require a large variety of skills, and to minimize the risk of \textsc{SkillExtractor} grouping two distinct skills as just one latent skill. 

\section{Impact of Training Time on Synthetic Student}
\begin{figure}[H]
    \centering
    \includegraphics[width=0.5\linewidth]{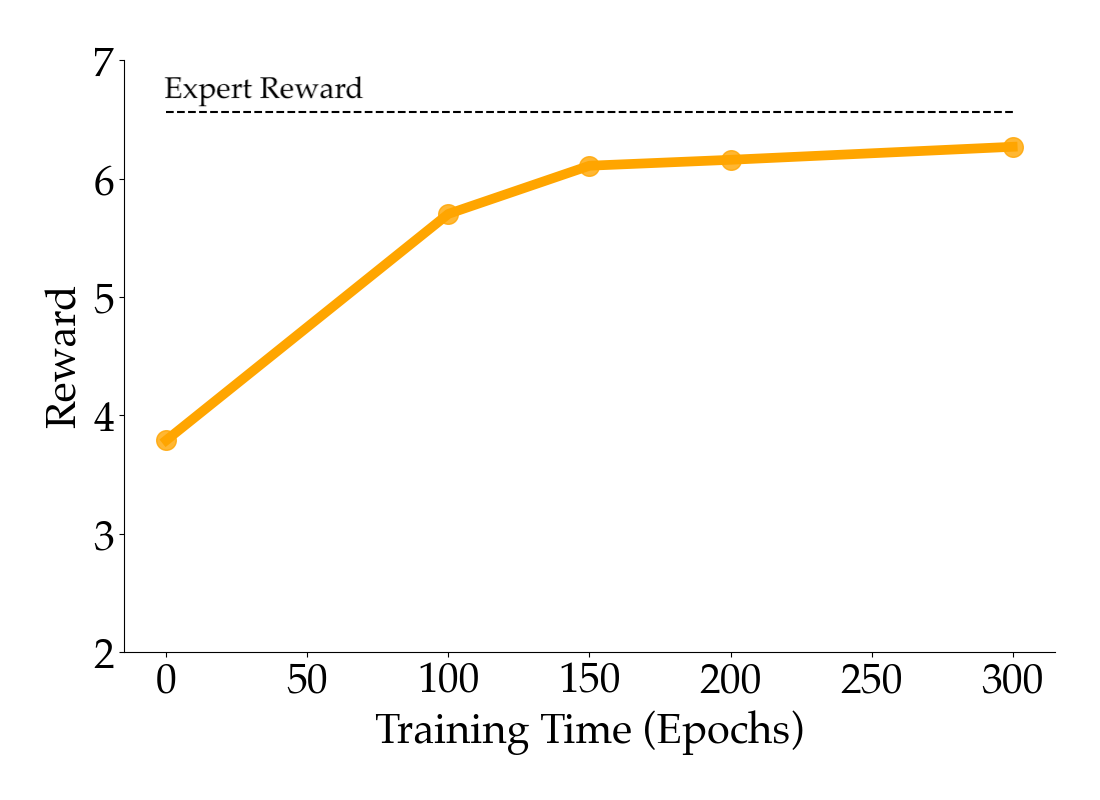}
    \caption{Reward starts to plateau over training iterations for the "reversing difficulty" synthetic student. Reported values are average reward over 100 random rollouts.}
    \label{fig:synthetic_parking_time}
\end{figure}

Although we report results for both half-trained and reverse difficulty synthetic students after finetuning on 100 epochs, one natural question is the effect of training time. We examine this more closely with the "reversing difficulty" student, where Fig. \ref{fig:synthetic_parking_time} shows that as we increase the number of training epochs (equivalent to adjusting the $N_{rep}$ parameter in our IL setting) for a fixed set of 3 drills, student reward starts to plateau close to the average reward the expert receives. This shows that for our synthetic model, the largest learning gain occurs at the start of training.  

In practice, a teacher could also adjust the $N_{drills}$ parameter and repeat the entire teaching session by repeating the expertise identification method in Alg. \ref{alg:expertise_identification}, which may return different skills as a student improves over time. Planning the optimal overall curriculum is likely extremely task-dependent, and requires a much stronger model of student learning that incorporates concepts such as forgetting, which we leave for future work.

\section{Source Code}
We provide all source code necessary to replicate our user study, for both \textsc{Parking} and \textsc{Writing} environments, as well as the trained CompILE models for both environments, at \url{https://github.com/Stanford-ILIAD/teaching}. 

\section{Environment Pre-Processing}
As described in Sec. 5 of the main paper, our \textsc{Parking} environment is built off of the HighwayEnv goal-based task whereas our \textsc{Writing} environment is custom built based off of the Omniglot dataset. For the purpose of simplifying our user study, we make the following modifications:

\begin{enumerate}[nosep,leftmargin=*]
    \item For \textsc{Parking}, while we train our expert agent and skill-discovery algorithms across all possible parking goals, we only pick goals in the bottom right quadrant for teaching students in our user study as a simplification. To expand to all goals in the task, we believe further teaching time would be necessary due to the larger number and variety of skills required. 
    \item Likewise, for \textsc{Writing}, we limit our sequences to contain only up to 5 different Balinese characters (``Na'', ``Ma'', ``Pa'', ``Ba'', ``Wa'') to reduce the amount of skills required to learn the overall task.
    \item Because crowdworkers in the Omniglot dataset differ in terms of interfaces used at the time of data collection, we ``infill'' all action sequences as a method of standardization. Specifically, we infill between any two consecutive states that differ by more than 1 pixel. Because these infilled trajectories are used to train our CompILE module for \textsc{SkillExtractor}, we likewise infill all user trajectories collected in our user study.  
\end{enumerate}

\section{Hyperparameters \& Training Details}
Here, we describe all necessary hyperparameters  to replicate training our (i) \textsc{Parking} expert agent, (ii) skill-discovery CompILE modules for both \textsc{Parking} and \textsc{Writing} tasks, and (iii) synthetic students for \textsc{Parking}. All models are trained on 1 NVIDIA TITAN RTX GPU, and the longest training time (for the expert \textsc{Parking} agent) is roughly 5 hours. 

\begin{enumerate}[nosep,leftmargin=*]
    \item \textbf{\textsc{Parking} expert agent}: We train a StableBaselines3 implementation of Soft Actor-Critic for $10^6$ epochs with a learning rate of 0.001, which achieves a roughly 100\% parking success rate. 
    \item \textbf{\textsc{Parking} CompILE module}: We train a  CompILE module (using the code from \cite{kipf2019compile}) for 2000 iterations with a learning rate of 0.001, batch size of 100, latent dimension of 16 (i.e. 16 possible skills), prior expected length of skill segments of 10, and prior number of segments per expert demonstration of 4. As described in the main paper, for \textsc{Parking}, we add a penalty to the original loss function that is the MSE loss between the state differences between two consecutive states in the CompILE reconstruction and that in the training data.  
    \item \textbf{\textsc{Writing} CompILE module}: We train a  CompILE module for 80 iterations with a learning rate of 0.005, batch size of 50,  latent dimension of 24 (i.e. 24 possible skills), prior expected length of skill segments of 250, and prior number of segments per expert demonstration of 8. For both \textsc{Parking} and \textsc{Writing}, we find it necessary to set the latent dimension high as many latent codes correspond to zero skill segments.  
    \item \textbf{\textsc{Parking} synthetic ``half-trained'' student}: We train a 4-layer feed-forward neural network for 50 epochs with a learning rate of 0.0005 and batch size 256 via behavior cloning on rollouts from the expert agent, which receives an eval MSE loss of 0.049. 
    \item \textbf{\textsc{Parking} synthetic ``reversing difficulty'' student}: We train a 4-layer feed-forward neural network for 400 epochs with a learning rate of 0.0005 and batch size 256 via behavior cloning on rollouts from the expert agent with only 20\% of the data containing reverse acceleration actions (negative y-value), which receives an eval MSE loss of 0.021.
\end{enumerate}

\section{User Study}
We recruited users on Prolific, a crowdsourcing platform to conduct research studies, as part of an IRB-approved study (Protocol No. 49406 reviewed by Stanford University). We recruited up to 25 users for each setting in our user study for both \textsc{Writing} and \textsc{Parking} tasks. Overall, participants were paid an estimated wage of 15 dollars per hour, and took on average 20 minutes to complete the entire study, including reading instructions, learning the motor control task, and completing a post-task survey. 

Each student was provided a link to an instructions page, where they provided a username to access the interfaces we built for both tasks. Each interface included step-by-step instructions on the side. As described in the main paper, students participated in a series of pre-test tries at the task, a sequence of practice sessions, and then an evaluation round. In the \textsc{Parking} task, due to its difficulty, practice sessions consisted of both expert demonstrations (with joystick movement corresponding to expert actions) and student practice mode, while the \textsc{Writing} task practice sessions consisted only of practice (of either full sequences, skills, or drills). Furthermore, to help guide students, we overlayed the state sequence of the target skill/drill/full-trajectory during demo and practice sessions for all settings in \textsc{Parking}. Finally, for both motor control tasks we imposed a time limit on students for pre-test, evaluation, and practice sessions, proportional to the length of the sequence. For fair comparison, we ensure that the total number of allowed time-steps is roughly equivalent between settings we directly compare with each other. We include images of the instructions and example interfaces for both tasks below.

Finally, each student participant completed a post-task survey where they provided ratings for how helpful they found the practice sessions for learning the control task, information about whether students used a trackpad or computer mouse, as well as any feedback about the interface or task itself. We found no significant impact on performance from whether a students used a trackpad or computer mouse. We provide the complete list of survey questions asked below. 

Overall, students found the \textsc{Parking} task particularly challenging, often asking for more practice sessions, which many found helpful (e.g. \textit{``It was interesting to give it a go and see how I improved in a short time.''}, \textit{``I could see that I was improving as the experiment went on''}). Meanwhile, students participating in the \textsc{Writing} task students enjoyed the educational experience of learning a new script  (e.g. \textit{``I am interested in writing forms and would one day like to learn some unusual scripts.''}, \textit{``interesting learning to write another language''}), but wished to learn more about the characters' meaning, motivating further research in making automatically-discovered skills (which may not necessarily be characters) more interpretable to students (e.g. \textit{``I would like to know what Balinese characters I'm tracing and their meaning''}). 
\begin{figure}[H]
    \centering
    \includegraphics[width=0.9\linewidth]{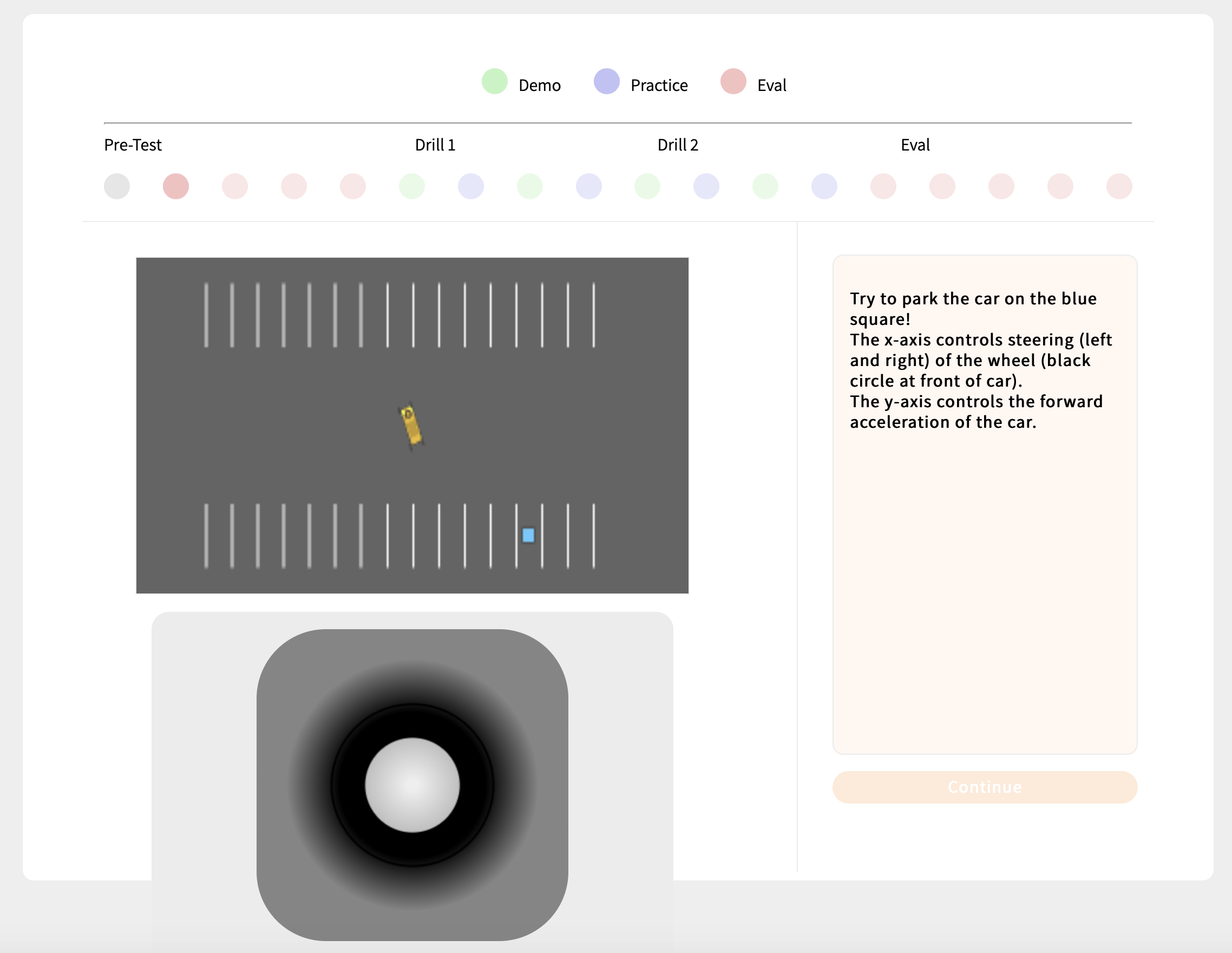}
    \caption{User study interface for the \textsc{Parking} task where student participants learn how to park a car with a joystick controller. A black circle marks the front of the car, and the blue square marks the goal parking spot.}
    \label{fig:parking_interface}
\end{figure}
\begin{figure}[H]
    \centering
    \includegraphics[width=0.9\linewidth]{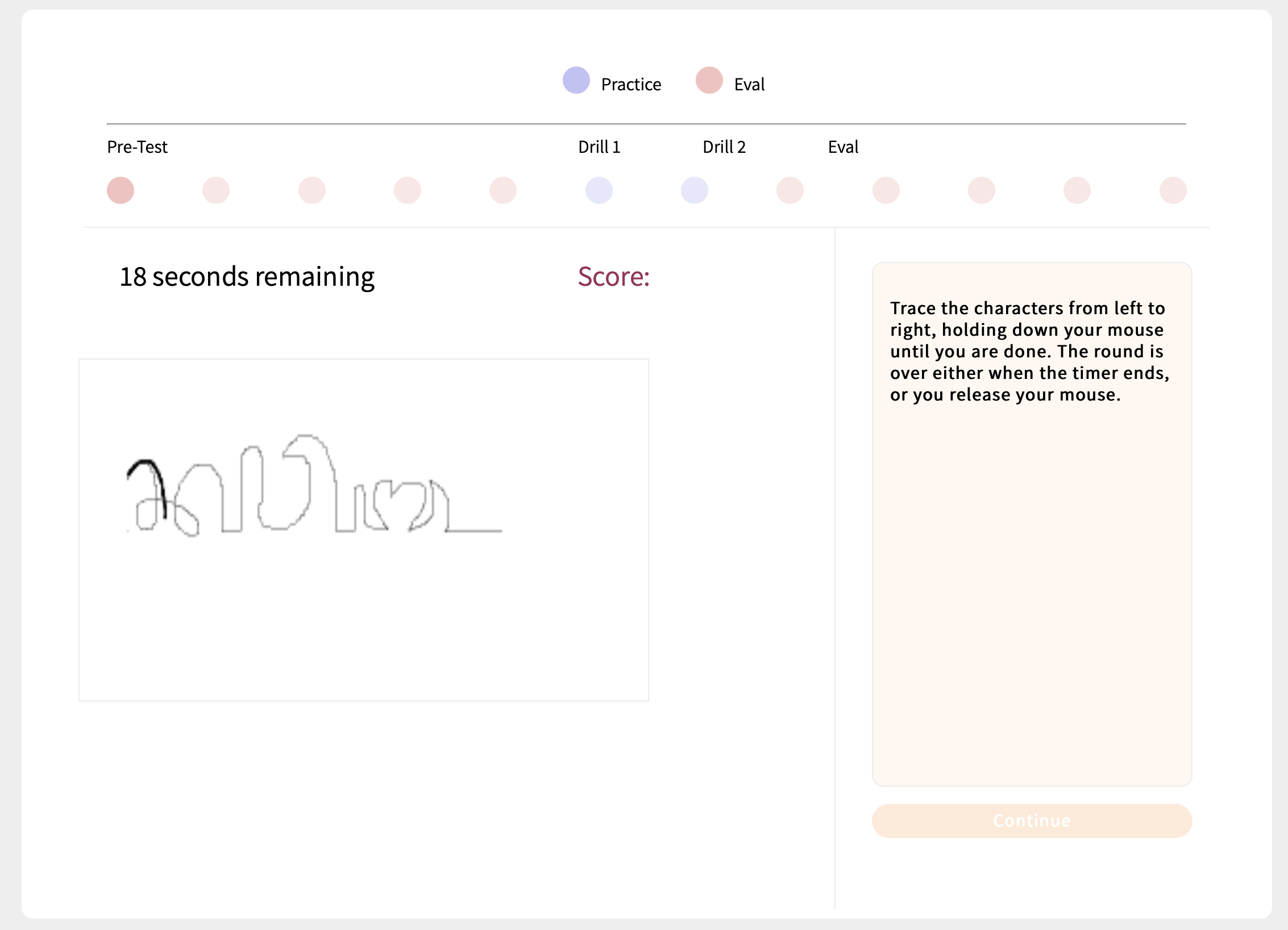}
    \caption{User study interface for the \textsc{Writing} task where student participants learn how to trace Balinese characters. After the user lets go of their mouse, or when the timer is over, a score representing the reward would be displayed.}
    \label{fig:writing_interface}
\end{figure}

\begin{figure}[H]
    \includegraphics[width=\linewidth]{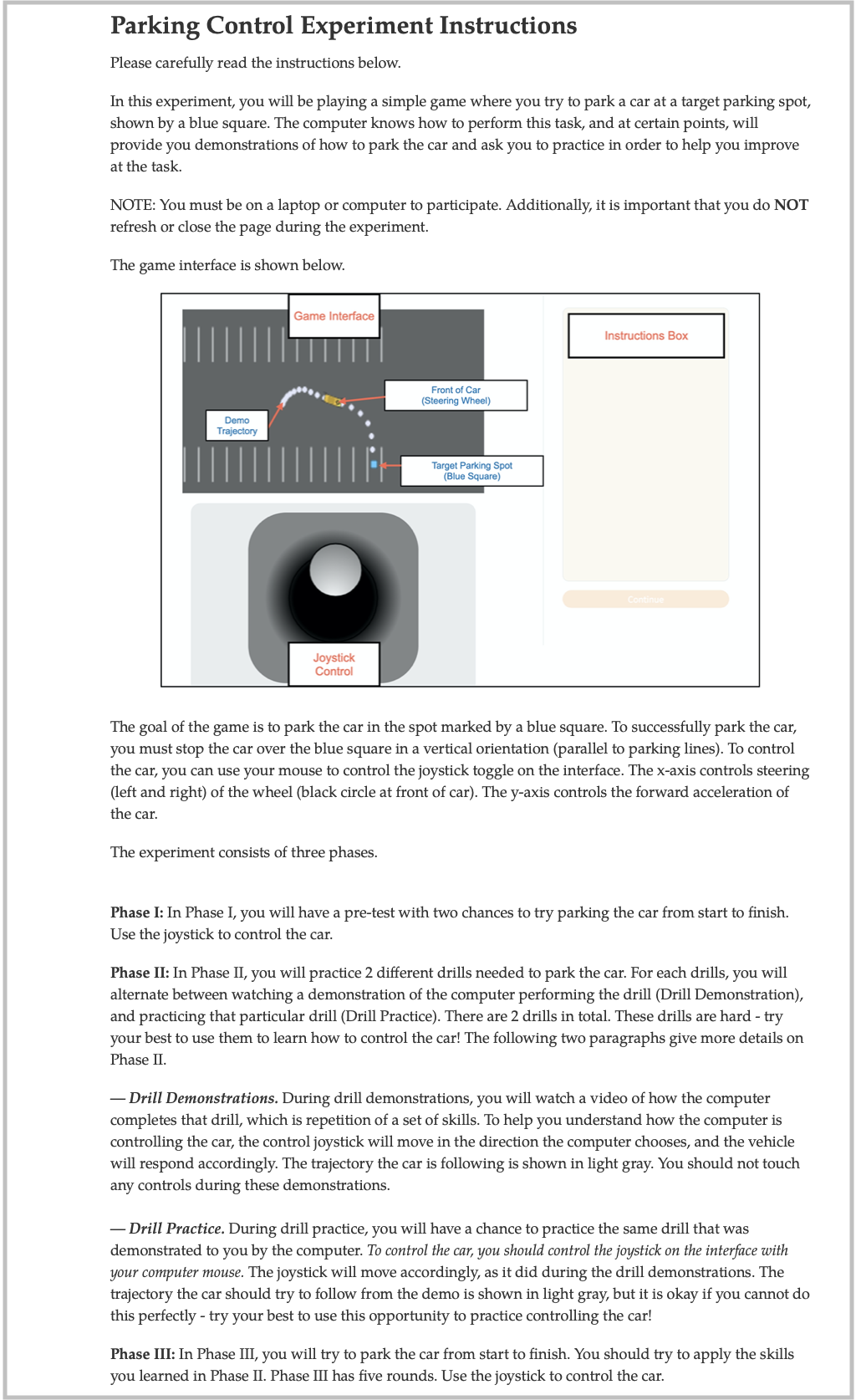}
    \caption{Instructions for the \textsc{Parking} task where student participants learn how to park a car with a joystick controller.}
    \label{fig:parking_instructions}
\end{figure}

\begin{figure}[H]
    \centering
    \includegraphics[width=\linewidth]{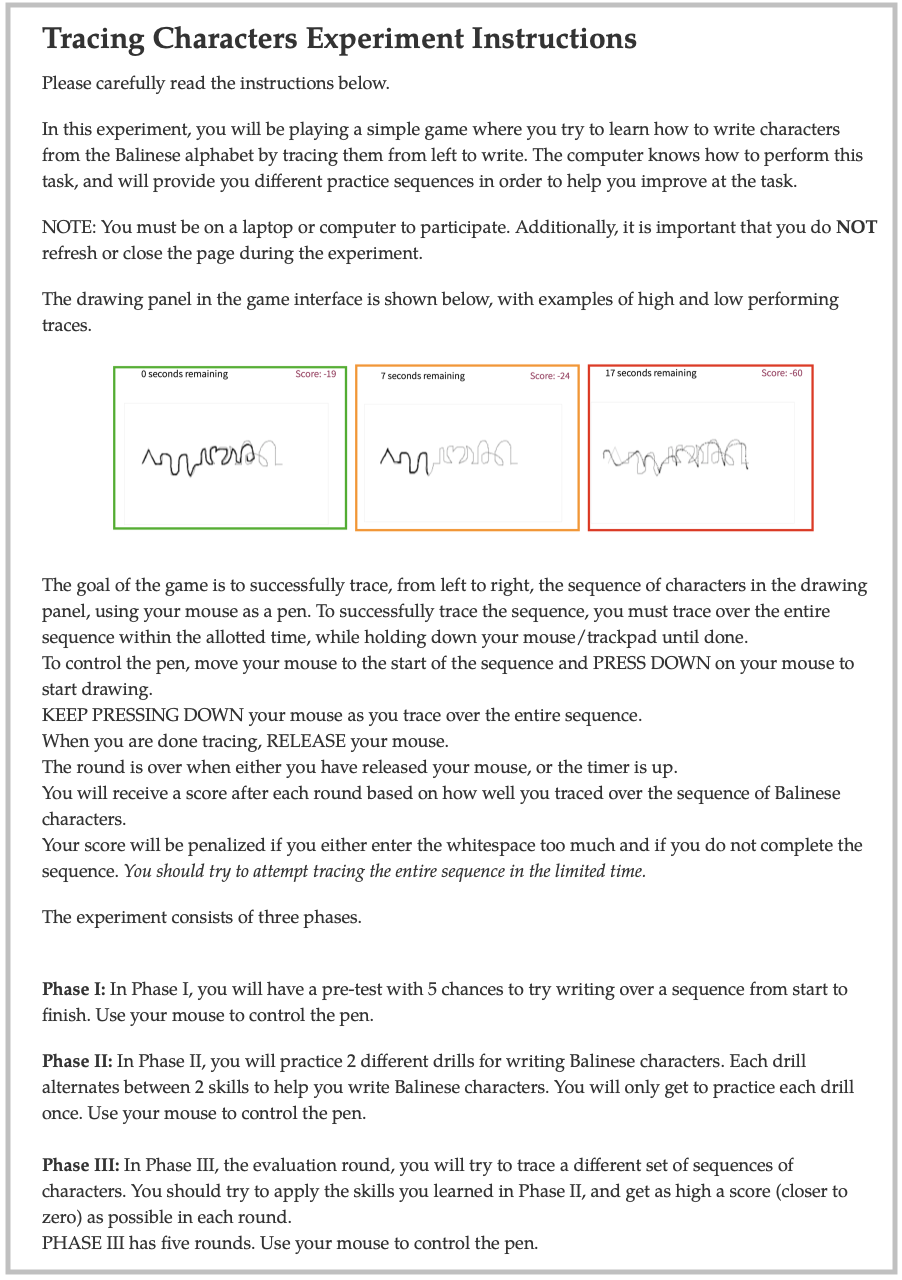}
    \caption{Instructions for the \textsc{Writing} task where student participants learn how to trace Balinese characters.}
    \label{fig:writing_instructions}
\end{figure}

\newpage

\textbf{User Study Survey Questions:}
\begin{enumerate}[nosep,leftmargin=*]
    \item Did you find the demos and practice sessions useful in learning how to park the car? / Did you find the practice sessions helpful in learning how to write the different characters? (Rating 1-7) 
    \item How easy was it to learn how to park the car? / How easy was it to learn how to write the characters? (Rating 1-7) 
    \item How easy was it to learn how to park the car? / How easy was it to learn how to write the characters? (Rating 1-7) 
    \item What else would have been helpful to learn how to park the car? / What else would have been helpful to learn how to write the characters?  
    \item What did you like about the experiment?  
    \item What would you wish to change about the experiment? 
    \item Did you use a laptop trackpad or a mouse to complete this study? 
\end{enumerate}

%% file: main.bbl
\begin{thebibliography}{44}
\providecommand{\natexlab}[1]{#1}
\providecommand{\url}[1]{\texttt{#1}}
\expandafter\ifx\csname urlstyle\endcsname\relax
  \providecommand{\doi}[1]{doi: #1}\else
  \providecommand{\doi}{doi: \begingroup \urlstyle{rm}\Url}\fi

\bibitem[Javdani et~al.(2018)Javdani, Admoni, Pellegrinelli, Srinivasa, and
  Bagnell]{javdani2018shared}
Shervin Javdani, Henny Admoni, Stefania Pellegrinelli, Siddhartha~S Srinivasa,
  and J~Andrew Bagnell.
\newblock Shared autonomy via hindsight optimization for teleoperation and
  teaming.
\newblock \emph{The International Journal of Robotics Research (IJRR)},
  37\penalty0 (7):\penalty0 717--742, 2018.

\bibitem[Jeon et~al.(2020)Jeon, Losey, and Sadigh]{jeon2020shared}
Hong~Jun Jeon, Dylan~P Losey, and Dorsa Sadigh.
\newblock Shared autonomy with learned latent actions.
\newblock \emph{Robotics: Science and Systems (RSS)}, 2020.

\bibitem[Losey et~al.(2021)Losey, Jeon, Li, Srinivasan, Mandlekar, Garg, Bohg,
  and Sadigh]{losey2021learning}
Dylan Losey, Hong~Jun Jeon, Mengxi Li, Krishnan Srinivasan, Ajay Mandlekar,
  Animesh Garg, Jeannette Bohg, and Dorsa Sadigh.
\newblock Learning latent actions to control assistive robots.
\newblock In \emph{Journal of Autonomous Robots (AURO)}, 2021.

\bibitem[Reddy et~al.(2018)Reddy, Dragan, and Levine]{reddy2018shared}
Siddharth Reddy, Anca~D Dragan, and Sergey Levine.
\newblock Shared autonomy via deep reinforcement learning.
\newblock \emph{Robotics: Science and Systems (RSS)}, 2018.

\bibitem[Karamcheti et~al.(2021)Karamcheti, Srivastava, Liang, and
  Sadigh]{karamcheti2021lila}
Siddharth Karamcheti, Megha Srivastava, Percy Liang, and Dorsa Sadigh.
\newblock {LILA}: Language-informed latent actions.
\newblock In \emph{Conference on Robot Learning (CoRL)}, 2021.

\bibitem[Rafferty et~al.(2016)Rafferty, Brunskill, Griffiths, and
  Shafto]{rafferty2016faster}
Anna~N. Rafferty, Emma Brunskill, Thomas~L. Griffiths, and Patrick Shafto.
\newblock Faster teaching via pomdp planning.
\newblock \emph{Cognitive Science}, 40 6:\penalty0 1290--332, 2016.

\bibitem[Bassen et~al.(2020)Bassen, Balaji, Schaarschmidt, Thille, Painter,
  Zimmaro, Games, Fast, and Mitchell]{bassen2020reinforcement}
Jonathan Bassen, Bharathan Balaji, Michael Schaarschmidt, Candace Thille, Jay
  Painter, Dawn Zimmaro, Alex Games, Ethan Fast, and John~C Mitchell.
\newblock Reinforcement learning for the adaptive scheduling of educational
  activities.
\newblock \emph{Conference on Human Factors in Computing Systems (CHI)}, 2020.

\bibitem[Poesia et~al.(2021)Poesia, Dong, and Goodman]{poesia2021contrastive}
Gabriel Poesia, WenXin Dong, and Noah Goodman.
\newblock Contrastive reinforcement learning of symbolic reasoning domains.
\newblock \emph{Advances in Neural Information Processing Systems (NeurIPS)},
  2021.

\bibitem[Srivastava and Goodman(2021)]{srivastava2021question}
Megha Srivastava and Noah Goodman.
\newblock Question generation for adaptive education.
\newblock \emph{Association for Computational Linguistics (ACL)}, 2021.

\bibitem[Mu et~al.(2021)Mu, Wang, Andersen, and Brunskill]{mu2021automatic}
Tong Mu, Shuhan Wang, Erik Andersen, and Emma Brunskill.
\newblock Automatic adaptive sequencing in a webgame.
\newblock \emph{International Conference on Intelligent Tutoring Systems},
  2021.

\bibitem[Agay(1982)]{agay1982teaching}
Denes Agay.
\newblock \emph{Teaching Piano: A Comprehensive Guide and Reference Book for
  the Instructor}.
\newblock Music Sales Amer, 1982.

\bibitem[Sutton et~al.(1999)Sutton, Precup, and Singh]{sutton1999between}
Richard~S Sutton, Doina Precup, and Satinder Singh.
\newblock Between mdps and semi-mdps: A framework for temporal abstraction in
  reinforcement learning.
\newblock \emph{Artificial Intelligence}, 112\penalty0 (1-2):\penalty0
  181--211, 1999.

\bibitem[Barto and Mahadevan(2003)]{barto2003recent}
Andrew~G Barto and Sridhar Mahadevan.
\newblock Recent advances in hierarchical reinforcement learning.
\newblock \emph{Discrete Event Dynamic Systems}, 13\penalty0 (1):\penalty0
  41--77, 2003.

\bibitem[McGovern and Sutton(1998)]{mcgovern1998macro}
Amy McGovern and Richard~S Sutton.
\newblock Macro-actions in reinforcement learning: An empirical analysis.
\newblock \emph{Computer Science Department Faculty Publication Series},
  page~15, 1998.

\bibitem[Parr and Russell(1997)]{parr1997reinforcement}
Ronald Parr and Stuart Russell.
\newblock Reinforcement learning with hierarchies of machines.
\newblock \emph{Advances in Neural Information Processing Systems (NIPS)},
  1997.

\bibitem[Dietterich(2000)]{dietterich2000hierarchical}
Thomas~G Dietterich.
\newblock Hierarchical reinforcement learning with the {MAXQ} value function
  decomposition.
\newblock \emph{Journal of Artificial Intelligence Research}, 13:\penalty0
  227--303, 2000.

\bibitem[Schaal et~al.(2005)Schaal, Peters, Nakanishi, and
  Ijspeert]{schaal2005learning}
Stefan Schaal, Jan Peters, Jun Nakanishi, and Auke Ijspeert.
\newblock Learning movement primitives.
\newblock \emph{Robotics Research. The Eleventh International Symposium}, pages
  561--572, 2005.

\bibitem[Daniel et~al.(2013)Daniel, Neumann, and Peters]{daniel2013autonomous}
Christian Daniel, Gerhard Neumann, and Jan Peters.
\newblock Autonomous reinforcement learning with hierarchical reps.
\newblock \emph{IEEE International Joint Conference on Neural Networks
  (IJCNN)}, 2013.

\bibitem[Ranchod et~al.(2015)Ranchod, Rosman, and
  Konidaris]{ranchod2015nonparametric}
Pravesh Ranchod, Benjamin Rosman, and George Konidaris.
\newblock Nonparametric bayesian reward segmentation for skill discovery using
  inverse reinforcement learning.
\newblock \emph{IEEE/RSJ International Conference on Intelligent Robots and
  Systems (IROS)}, 2015.

\bibitem[Bacon et~al.(2017)Bacon, Harb, and Precup]{bacon2017option}
Pierre-Luc Bacon, Jean Harb, and Doina Precup.
\newblock The option-critic architecture.
\newblock \emph{AAAI Conference on Artificial Intelligence}, 2017.

\bibitem[Bagaria and Konidaris(2019)]{bagaria2019option}
Akhil Bagaria and George Konidaris.
\newblock Option discovery using deep skill chaining.
\newblock \emph{International Conference on Learning Representations (ICLR)},
  2019.

\bibitem[Gregor et~al.(2016)Gregor, Rezende, and
  Wierstra]{gregor2016variational}
Karol Gregor, Danilo~Jimenez Rezende, and Daan Wierstra.
\newblock Variational intrinsic control.
\newblock \emph{arXiv preprint arXiv:1611.07507}, 2016.

\bibitem[Florensa et~al.(2017)Florensa, Duan, and
  Abbeel]{florensa2017stochastic}
Carlos Florensa, Yan Duan, and Pieter Abbeel.
\newblock Stochastic neural networks for hierarchical reinforcement learning.
\newblock \emph{International Conference on Learning Representations (ICLR)},
  2017.

\bibitem[Achiam et~al.(2018)Achiam, Edwards, Amodei, and
  Abbeel]{achiam2018variational}
Joshua Achiam, Harrison Edwards, Dario Amodei, and Pieter Abbeel.
\newblock Variational option discovery algorithms.
\newblock \emph{arXiv preprint arXiv:1807.10299}, 2018.

\bibitem[Eysenbach et~al.(2019)Eysenbach, Gupta, Ibarz, and
  Levine]{eysenbach2019diversity}
Benjamin Eysenbach, Abhishek Gupta, Julian Ibarz, and Sergey Levine.
\newblock Diversity is all you need: Learning skills without a reward function.
\newblock \emph{International Conference on Learning Representations (ICLR)},
  2019.

\bibitem[Sharma et~al.(2020)Sharma, Gu, Levine, Kumar, and
  Hausman]{sharma2020dynamics}
Archit Sharma, Shixiang Gu, Sergey Levine, Vikash Kumar, and Karol Hausman.
\newblock Dynamics-aware unsupervised discovery of skills.
\newblock \emph{International Conference on Learning Representations}, 2020.

\bibitem[Campos et~al.(2020)Campos, Trott, Xiong, Socher, Gir{\'o}-i Nieto, and
  Torres]{campos2020explore}
V{\'\i}ctor Campos, Alexander Trott, Caiming Xiong, Richard Socher, Xavier
  Gir{\'o}-i Nieto, and Jordi Torres.
\newblock Explore, discover and learn: Unsupervised discovery of state-covering
  skills.
\newblock \emph{International Conference on Machine Learning (ICML)}, 2020.

\bibitem[Frans et~al.(2018)Frans, Ho, Chen, Abbeel, and
  Schulman]{frans2018meta}
Kevin Frans, Jonathan Ho, Xi~Chen, Pieter Abbeel, and John Schulman.
\newblock Meta learning shared hierarchies.
\newblock \emph{International Conference on Learning Representations (ICLR)},
  2018.

\bibitem[Kipf et~al.(2019)Kipf, Li, Dai, Zambaldi, Sanchez-Gonzalez,
  Grefenstette, Kohli, and Battaglia]{kipf2019compile}
Thomas Kipf, Yujia Li, Hanjun Dai, Vinicius Zambaldi, Alvaro Sanchez-Gonzalez,
  Edward Grefenstette, Pushmeet Kohli, and Peter Battaglia.
\newblock {CompILE}: Compositional imitation learning and execution.
\newblock \emph{International Conference on Machine Learning (ICML)}, 2019.

\bibitem[Shankar et~al.(2019)Shankar, Tulsiani, Pinto, and
  Gupta]{shankar2019discovering}
Tanmay Shankar, Shubham Tulsiani, Lerrel Pinto, and Abhinav Gupta.
\newblock Discovering motor programs by recomposing demonstrations.
\newblock \emph{International Conference on Learning Representations (ICLR)},
  2019.

\bibitem[Dragan and Srinivasa(2013)]{dragan2013policy}
Anca~D Dragan and Siddhartha~S Srinivasa.
\newblock A policy-blending formalism for shared control.
\newblock \emph{The International Journal of Robotics Research (IJRR)},
  32\penalty0 (7):\penalty0 790--805, 2013.

\bibitem[Reddy et~al.(2021)Reddy, Levine, and Dragan]{reddy2021assisted}
Siddharth Reddy, Sergey Levine, and Anca Dragan.
\newblock Assisted perception: Optimizing observations to communicate state.
\newblock In \emph{Conference on Robot Learning}, pages 748--764. PMLR, 2021.

\bibitem[Bragg and Brunskill(2019)]{bragg2020fake}
Jonathan Bragg and Emma Brunskill.
\newblock Fake it till you make it: Learning-compatible performance support.
\newblock \emph{Conference on Uncertainty in Artificial Intelligence (UAI)},
  2019.

\bibitem[Corbett and Anderson(1994)]{corbett1994knowledge}
Albert~T Corbett and John~R Anderson.
\newblock Knowledge tracing: Modeling the acquisition of procedural knowledge.
\newblock \emph{User modeling and user-adapted interaction}, 4\penalty0
  (4):\penalty0 253--278, 1994.

\bibitem[Piech et~al.(2015)Piech, Bassen, Huang, Ganguli, Sahami, Guibas, and
  Sohl-Dickstein]{piech2015deep}
Chris Piech, Jonathan Bassen, Jonathan Huang, Surya Ganguli, Mehran Sahami,
  Leonidas~J Guibas, and Jascha Sohl-Dickstein.
\newblock Deep knowledge tracing.
\newblock \emph{Advances in Neural Information Processing Systems (NIPS)},
  2015.

\bibitem[Doroudi et~al.(2019)Doroudi, Aleven, and Brunskill]{doroudi2019where}
Shayan Doroudi, Vincent Aleven, and Emma Brunskill.
\newblock Where's the reward? a review of reinforcement learning for
  instructional sequencing.
\newblock \emph{International Journal of Artifical Intelligence in Education},
  29:\penalty0 568--620, 2019.

\bibitem[Zhu(2015)]{zhu2015machine}
Xiaojin Zhu.
\newblock Machine teaching: An inverse problem to machine learning and an
  approach toward optimal education.
\newblock \emph{AAAI Conference on Artificial Intelligence}, 2015.

\bibitem[Mac~Aodha et~al.(2018)Mac~Aodha, Su, Chen, Perona, and
  Yue]{mac2018teaching}
Oisin Mac~Aodha, Shihan Su, Yuxin Chen, Pietro Perona, and Yisong Yue.
\newblock Teaching categories to human learners with visual explanations.
\newblock In \emph{Proceedings of the IEEE Conference on Computer Vision and
  Pattern Recognition}, pages 3820--3828, 2018.

\bibitem[Krahe(1999)]{krahe1999motor}
Natalie Krahe.
\newblock \emph{Motor Skill Learning: A Review of Trends and Theories}.
\newblock Caroll College, 1999.

\bibitem[Shumway-Cook and Woollacott(1995)]{shumway1995theory}
Anne Shumway-Cook and Marjorie~H Woollacott.
\newblock \emph{Motor control: Theory and practical applications}.
\newblock Lippincqtt Williams \& Wilkins, 1995.

\bibitem[Rosenbaum(2010)]{rosenbaum2009human}
David~A Rosenbaum.
\newblock \emph{Human motor control}.
\newblock Elsevier, 2010.

\bibitem[Lake et~al.(2015)Lake, Salakhutdinov, and Tenenbaum]{lake2015human}
Brenden~M Lake, Ruslan Salakhutdinov, and Joshua~B Tenenbaum.
\newblock Human-level concept learning through probabilistic program induction.
\newblock \emph{Science}, 350\penalty0 (6266):\penalty0 1332--1338, 2015.

\bibitem[Leurent(2018)]{highway-env}
Edouard Leurent.
\newblock An environment for autonomous driving decision-making.
\newblock \url{https://github.com/eleurent/highway-env}, 2018.

\bibitem[Hill et~al.(2018)Hill, Raffin, Ernestus, Gleave, Kanervisto, Traore,
  Dhariwal, Hesse, Klimov, Nichol, Plappert, Radford, Schulman, Sidor, and
  Wu]{stable-baselines}
Ashley Hill, Antonin Raffin, Maximilian Ernestus, Adam Gleave, Anssi
  Kanervisto, Rene Traore, Prafulla Dhariwal, Christopher Hesse, Oleg Klimov,
  Alex Nichol, Matthias Plappert, Alec Radford, John Schulman, Szymon Sidor,
  and Yuhuai Wu.
\newblock Stable baselines.
\newblock \url{https://github.com/hill-a/stable-baselines}, 2018.

\end{thebibliography}
